\documentclass[11pt, a4paper, logo, copyright, nonumbering]{xiaomi}

\usepackage[numbers, sort&compress, square]{natbib}
\usepackage{dblfloatfix}
\usepackage[normalem]{ulem}
\usepackage{caption}
\usepackage{dramatist}
\usepackage{xspace}
\usepackage{pifont} % http://ctan.org/pkg/pifont
\usepackage{multirow}
\usepackage{tcolorbox}
\usepackage{xltabular}
\usepackage{longtable}
\usepackage{hyperref}
\usepackage{wrapfig}

\usepackage{amsfonts}
\usepackage{amsmath}
\usepackage{amssymb}
\usepackage{lineno}
\usepackage{multirow}
\usepackage{adjustbox}

\usepackage[bottom]{footmisc}

\usepackage{CJKutf8}
\usepackage{setspace}
\usepackage{makecell}
\usepackage{graphicx}
\usepackage{subcaption}
\usepackage{multicol} % contributors

\usepackage[utf8]{inputenc} % allow utf-8 input
\usepackage[T1]{fontenc}    % use 8-bit T1 fonts
\usepackage{hyperref}
\usepackage{cleveref}
\usepackage{url}            % simple URL typesetting
\usepackage{booktabs}       % professional-quality tables
\usepackage{amsfonts}       % blackboard math symbols
\usepackage{nicefrac}       % compact symbols for 1/2, etc.
\usepackage{microtype}      % microtypography
\usepackage{xcolor}         % colors
\usepackage{colortbl}
\usepackage{tcolorbox}
\usepackage{xspace}
\definecolor{BrickRed}{rgb}{.72,0,0}
\definecolor{darkgreen}{rgb}{0.0, 0.5, 0.0}
\definecolor{ForestGreen}{RGB}{34,139,34}
\definecolor{LakeBlue}{RGB}{0,61,153}
\definecolor{MiOrange}{RGB}{255,225,204}
\definecolor{Hex}{RGB}{225,213,231}

\hypersetup{
    colorlinks=true,
    allcolors=LakeBlue
}

\usepackage{amsmath,amssymb}
\usepackage{bm}
\usepackage{booktabs}
\usepackage{graphicx}
\usepackage{multirow}
\usepackage{hyperref}
\usepackage{fontawesome5}
\usepackage{url}
\usepackage[table]{xcolor}
\usepackage{pifont}
\newcommand{\yesanchor}{\ding{51}}
\newcommand{\noanchor}{\ding{55}}
\usepackage{soul}
\usepackage{float}
\usepackage{wrapfig}
\definecolor{linkblue}{RGB}{0,92,160}
\definecolor{tablehead}{RGB}{232,240,248}
\definecolor{tablegroup}{RGB}{246,248,250}
\definecolor{tableours}{RGB}{239,235,248}
\definecolor{starorange}{RGB}{230,170,20}
\definecolor{projectblue}{HTML}{2563EB}
\definecolor{codegray}{HTML}{334155}
\definecolor{datateal}{HTML}{0D9488}
\definecolor{modelpurple}{HTML}{7C3AED}

\newcommand{\resourceLink}[4]{%
	\href{#1}{%
		\mbox{%
			{\color{#2}\fontsize{12}{14}\selectfont #3}%
			\hspace{0.35em}%
			{\color{black}#4}%
		}%
	}%
}

\hypersetup{
	colorlinks=true,
	citecolor=linkblue,
	linkcolor=linkblue,
	urlcolor=linkblue
}

\definecolor{linkred}{RGB}{255,0,0}

\newcommand{\redfigref}[1]{%
	Fig.~{\setlength{\fboxsep}{1pt}%
		\setlength{\fboxrule}{0.8pt}%
		\fcolorbox{linkred}{white}{%
			\hypersetup{linkcolor=black}\ref{#1}}}%
}

\newcommand{\redsecref}[1]{%
	Sec.~{\setlength{\fboxsep}{1pt}%
		\setlength{\fboxrule}{0.8pt}%
		\fcolorbox{linkred}{white}{%
			\hypersetup{linkcolor=black}\ref{#1}}}%
}

\newcommand{\redappref}[1]{%
	Appendix~{\setlength{\fboxsep}{1pt}%
		\setlength{\fboxrule}{0.8pt}%
		\fcolorbox{linkred}{white}{%
			\hypersetup{linkcolor=black}\ref{#1}}}%
}

\newcommand{\redeqref}[1]{%
	Eq.~{\setlength{\fboxsep}{1pt}%
		\setlength{\fboxrule}{0.8pt}%
		\fcolorbox{linkred}{white}{%
			\hypersetup{linkcolor=black}(\ref{#1})}}%
}

\newcommand{\redtabref}[1]{%
	Table~{\setlength{\fboxsep}{1pt}%
		\setlength{\fboxrule}{0.8pt}%
		\fcolorbox{linkred}{white}{%
			\hypersetup{linkcolor=black}\ref{#1}}}%
}

\newcommand{\method}{\textbf{WorldCast}}
\newcommand{\bench}{\textbf{CLACT}}
\newcommand{\benchmarkSentence}[1]{\sethlcolor{yellow!10}\hl{#1}}
\newcommand{\methodSentence}[1]{\sethlcolor{red!10}\hl{#1}}

\definecolor{linkpink}{RGB}{244,42,175}

\title{
	\centering
	\color{black}\bfseries
	WorldCast: World-Consistent Character\\
	Video Generation with Large-Angle Camera Control
}

\titlerunning{WorldCast}

\author{
	\shortstack[c]{
		\normalfont\fontsize{11}{14}\selectfont\color{black}
		Lei Cao\textsuperscript{1,2*},\enspace
		Yu Liu\textsuperscript{2*},\enspace
		Wenhua Liao\textsuperscript{2},\enspace
		Peng Zhang\textsuperscript{2$\ddagger$},\enspace
		Fei Wang\textsuperscript{2},\enspace
		\\
		\normalfont\fontsize{11}{14}\selectfont\color{black}
		Hao Zhang\textsuperscript{1},\enspace
		Jiajun Le\textsuperscript{1},\enspace
		Daiguo Zhou\textsuperscript{2$\dagger$},\enspace
		Jian Luan\textsuperscript{2},\enspace
		Jiayi Ma\textsuperscript{1,3$\dagger$}
	}
}

\institute{
	\begingroup
	\normalfont\fontsize{10}{13}\selectfont\color{black}
	\begin{tabular}{@{}c@{}}
		\textbf{\textsuperscript{1}Electronic Information School,
			Wuhan University}\quad
		\textbf{\textsuperscript{2}MiLM Plus, Xiaomi Inc.}\\[2pt]
		\textbf{\textsuperscript{3}School of Robotics,
			Wuhan University}\\[9pt]
		{\normalfont\fontsize{11}{14}\selectfont
			\resourceLink
			{https://pixctrol.github.io/WorldCast/}
			{projectblue}{\faGlobe}{Project Page}
			\quad
			\resourceLink
			{https://github.com/PixCtrol/WorldCast}
			{codegray}{\faGithub}{Code}
			\quad
			\resourceLink
			{https://huggingface.co/datasets/ly56/CLACT}
			{datateal}{\faDatabase}{Dataset}
			\quad
			\resourceLink
			{https://huggingface.co/ly56/worldcast}
			{modelpurple}{\faIcon{file-alt}}{Model Card}
		}
	\end{tabular}
	\endgroup
}

\begin{document}

\begin{abstract}
	Precise camera control is essential for cinematic character video generation, yet existing methods struggle to follow large-angle trajectories with combined rotation, translation, and radius variation from arbitrary initial azimuths.
	Such viewpoint changes also reveal character regions unseen in the initial frame, causing geometric distortions, appearance degradation, and identity drift.
	To systematically study these coupled challenges, we introduce \textit{\bench{}, \benchmarkSentence{a large-scale \textbf{C}haracter-centric dataset and benchmark covering diverse \textbf{L}arge-\textbf{A}ngle \textbf{C}amera \textbf{T}rajectories}}, with accurate camera geometry and multi-view character observations for training and evaluation.
	We further propose \textit{\method{}, \methodSentence{a framework for world-consistent character video generation under diverse large-angle camera motions}}.
	Specifically, we back-project the initial frame into a 3D point cloud using estimated depth, align the point-cloud scale with the target camera trajectory, and render it along the trajectory to obtain explicit geometric guidance and coverage masks.
	To complement the incomplete character information, we introduce Geometry-Aligned Memory Attention (GAMA).
	Using relative camera poses estimated jointly from the initial frame and multi-view character anchors, we express anchor and target-view geometry in a common coordinate frame.
	GAMA uses this geometric alignment to guide the retrieval and injection of view-relevant character features during generation.
	Evaluations across diverse large-angle camera trajectories demonstrate accurate trajectory following while preserving character geometry, appearance, and identity.

\end{abstract}

\maketitle

\begingroup
\renewcommand{\thefootnote}{}
\footnotetext{%
	\normalfont\fontsize{9}{11}\selectfont
	\textsuperscript{*}Equal Contribution.
	\quad
	\textsuperscript{$\dagger$}Co-corresponding Authors.
	\quad
	\textsuperscript{$\ddagger$}Project Lead.
}
\endgroup

\begin{figure}[t] 
	\vspace{-18pt}
	\centering 
	
	\includegraphics[width=\linewidth]{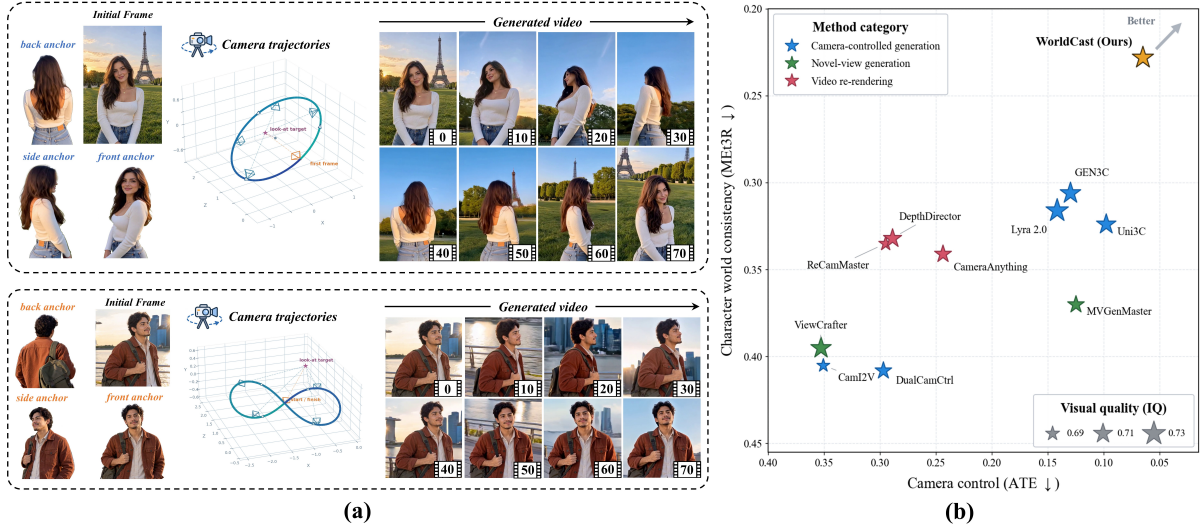} 
	\vspace{-18pt}
	\caption{ 
		\textbf{\textit{WorldCast} enables accurate large-angle camera control 
			while preserving character consistency.} 
		\textbf{(a)} Generated sequences along tilted circular 
		and lemniscate camera trajectories. 
		\textbf{(b)} Quantitative comparison of camera-control accuracy 
		(ATE $\downarrow$), cross-view geometric consistency 
		(MEt3R $\downarrow$), and image quality (IQ $\uparrow$). 
		Each star denotes a method; colors distinguish method categories, 
		while star size indicates image quality. 
		Both axes are reversed, such that methods with lower ATE and MEt3R 
		appear toward the upper right. 
	} 
	
	\label{fig:teaser} 
	\vspace{7pt} 
\end{figure}

\section{Introduction}

\label{sec:intro}

Recent video diffusion models have achieved remarkable visual fidelity in
text- and image-conditioned video generation~\citep{wan2025wan,yang2025cogvideox}.
For cinematic content creation, however, visual quality alone is insufficient:
creators also require precise control over camera motion~\citep{courant2024exceptional}.
In character-centric videos, camera motion can reveal a subject from different
viewpoints, establish spatial relationships, and direct visual attention.
Realizing such shots therefore requires not only accurate trajectory following,
but also consistent preservation of character geometry, appearance, and identity
across substantial viewpoint changes~\citep{yang2026gloria}.

Camera-controlled video generation has progressed from camera-pose and ray-based
conditioning to explicit geometric guidance derived from 3D renderings
~\citep{wang2024motionctrl,zheng2024cami2v,he2025cameractrl,
	ren2025gen3c,cao2025uni3c}.
Recent methods further explore joint appearance-depth modeling and more expressive
camera representations to support complex camera motions
~\citep{zhang2026unified,zhang2025dualcamctrl}.
Meanwhile, multi-view generation and video re-rendering provide complementary
means of improving cross-view consistency and synthesizing novel viewpoints
~\citep{zhou2025stable,cao2025mvgenmaster,bai2025recammaster,
	li2026cameraanything}.
Despite this progress, accurate trajectory following and
consistent character generation remain difficult under
large viewpoint changes.

Large-angle camera motion introduces two closely coupled challenges.
First, trajectories involving substantial rotation together with translation,
radius variation, and arbitrary initial azimuths place significantly stronger
demands on camera control.
Limited trajectory coverage or inaccurate geometric conditioning can result in
under-rotation, incorrect translation, and drift from the requested path
~\citep{yang2026confctrl}.
Second, the same viewpoint changes progressively expose character regions that
are unobserved in the initial frame, leaving their geometry, appearance, and
identity weakly constrained
~\citep{yu2024viewcrafter,Wu_2026_CVPR}.
These challenges are inherently linked: the target camera trajectory determines
both the geometry that the generated video must follow and the previously unseen
character content that must be synthesized at each viewpoint.
Explicit geometric guidance can constrain the former, while multi-view character
observations provide complementary evidence for the latter.
Effectively combining them therefore requires relating both sources of information
to the target viewpoints within a shared geometric reference frame.

Existing data remain insufficient for systematically studying and evaluating
large-angle character-centric camera control.
Current camera-motion datasets provide limited coverage of complex trajectories
involving substantial rotation, translation, radius variation, and diverse
initial azimuths~\citep{li2025realcam,bai2025recammaster}.
Moreover, camera-motion sequences are not consistently paired
with complementary views of the same character, limiting
the joint study of trajectory accuracy and cross-view
character consistency~\citep{Wu_2026_CVPR,wang2026harmoview}.
To address these limitations, we construct \textit{\bench{}}, a character-centric
dataset and benchmark designed for diverse large-angle camera motion.
Its training corpus contains $50{,}000$ camera-motion sequences rendered in
Unreal Engine 5 and $30{,}000$ Blender-rendered samples with multi-view
character anchors, while the evaluation set covers $42$ synthetic and real-world
character scenes under multiple trajectory families.
Together, these data support both training and systematic evaluation of
trajectory accuracy, character consistency, and visual quality across diverse
camera motions and substantial viewpoint changes.

Building on this benchmark, we propose \textit{\method{}},
a framework that couples scale-aligned geometric guidance
with geometry-aligned multi-view character memory.
We align the relative scale of initial-view depth and camera
translation, then back-project the initial frame into a
3D point cloud and render it along the target trajectory
to obtain geometric guidance and coverage masks.
To complement character information missing from the initial
view, we introduce Geometry-Aligned Memory Attention (GAMA).
Using camera poses jointly estimated from the initial frame
and character anchors, GAMA aligns anchor and target-view
geometry in a shared coordinate frame, enabling geometry-aware
retrieval of view-relevant character features and their
injection into the video tokens.
As shown in \redfigref{fig:teaser}, the framework improves
camera-control accuracy and cross-view character consistency
while maintaining competitive visual quality.

\textbf{Our contributions are threefold:}
\begin{itemize}
	\item We introduce \textit{\bench{}}, a character-centric dataset
	and benchmark combining diverse large-angle camera
	trajectories and multi-view character observations
	to support training and evaluation of camera control
	and character consistency.
	
	\item We propose \textit{\method{}}, a video generation framework
	that couples scale-aligned geometric guidance with
	multi-view character memory for accurate large-angle
	camera control and consistent character generation.
	
	\item We introduce Geometry-Aligned Memory Attention
	(GAMA), which aligns multi-view anchor geometry with
	the target camera trajectory in a shared coordinate
	frame, enabling geometry-aware retrieval and injection
	of view-relevant character features.
\end{itemize}

\section{Related Work}
\label{sec:related}

\paragraph{Camera-controllable video generation.}
Camera-controlled video generation incorporates explicit motion conditions into video diffusion models.
MotionCtrl disentangles camera and object motion, while CameraCtrl introduces camera conditioning
~\citep{wang2024motionctrl,he2025cameractrl}. CamCo and CamI2V incorporate camera geometry and epipolar
constraints into cross-frame modeling~\citep{xu2024camco,zheng2024cami2v}. VD3D and AC3D extend camera conditioning to video diffusion transformers~\citep{bahmani2025vd3d,bahmani2025ac3d},
while UCPE unifies camera pose, intrinsics, distortion, and orientation in a common representation
~\citep{zhang2026unified}. DualCamCtrl couples RGB and depth generation to strengthen
adherence to camera conditions~\citep{zhang2025dualcamctrl}. For video re-rendering, ReCamMaster, DepthDirector, and CameraAnything generate new camera trajectories while
preserving source content and dynamics~\citep{bai2025recammaster,chen2026beyond,li2026cameraanything}.

\paragraph{Geometry-aware and multi-view generation.}
Geometric representations and complementary observations provide spatial constraints for viewpoint synthesis.
3DiM, GeNVS, and CAT3D study novel-view synthesis from one or a few observations~\citep{watson2023novel,chan2023generative,gao2024cat3d}, while ViewCrafter, MVGenMaster, and SEVA leverage geometry or multiple input views to synthesize target viewpoints~\citep{yu2024viewcrafter,cao2025mvgenmaster,zhou2025stable}.
For camera-controlled video generation, RealCam-I2V, Uni3C, and GEN3C introduce explicit scene geometry or trajectory-aligned 3D guidance~\citep{li2025realcam,cao2025uni3c,ren2025gen3c}.
ConfCtrl improves robustness to unreliable projections~\citep{yang2026confctrl}, while Lyra~2.0 and WorldStereo further explore persistent geometry and geometry-constrained memory retrieval~\citep{shen2026lyra,zhang2026worldstereo}.
Our work instead relates multi-view character-anchor geometry to the target trajectory, enabling view-relevant character evidence to guide generation.

\paragraph{Identity-consistent video generation.}
Subject-conditioned video generation seeks to preserve appearance and identity across generated frames. VideoBooth and DreamVideo introduce subject-specific conditioning~\citep{jiang2024videobooth,wei2024dream},
while ConsisID and Magic-Me develop mechanisms for identity preservation~\citep{yuan2025identity,ma2024magic}.
ConsID-Gen uses complementary views to improve cross-view consistency, and Gloria represents character appearance through content anchors~\citep{Wu_2026_CVPR,yang2026gloria}. Virtually Being combines identity preservation across viewpoints with camera control through customization based on performance captures~\citep{xu2025virtually}. We jointly address character consistency and large-angle camera control through geometric guidance and view-aware
retrieval from multi-view character memory.

\section{Method}
\subsection{Preliminaries: Video Diffusion Models} 
Latent video diffusion models~\citep{blattmann2023align} operate in the compressed space of a pretrained variational autoencoder (VAE). Given a video $\bm{x}$, the encoder $\mathcal{E}$ produces a latent representation $\bm{z}_0=\mathcal{E}(\bm{x})\in\mathbb{R}^{d}$. Following the flow matching formulation~\citep{lipman2022flow}, we independently sample $\bm{z}_1\sim\mathcal{N}(\bm{0},\bm{I}_d)$ and define the linear interpolation: 
\begin{equation} 
	\bm{z}_{\tau} 
	= 
	(1-\tau)\bm{z}_0+\tau\bm{z}_1, 
	\qquad 
	\tau\in[0,1]. 
	\label{eq:flow_path} 
\end{equation} 

Conditioned on text and initial-frame features $c=(c_{\mathrm{txt}},c_{\mathrm{img}})$, a diffusion transformer $v_{\theta}$ learns the conditional velocity field with target velocity 
$\bm{u}_{\tau} 
= 
\frac{\mathrm{d}\bm{z}_{\tau}}{\mathrm{d}\tau} 
= 
\bm{z}_1-\bm{z}_0$ 
by minimizing: 
\begin{equation} 
	\mathcal{L}_{\mathrm{FM}}(\theta) 
	= 
	\mathbb{E}_{(\bm{z}_0,c)\sim p_{\mathrm{data}},\,\tau\sim p_{\tau},\,\bm{z}_1\sim\mathcal{N}(\bm{0},\bm{I}_d)} 
	\left[ 
	\left\| 
	v_{\theta}(\bm{z}_{\tau},\tau,c) 
	- 
	\bm{u}_{\tau} 
	\right\|_2^2 
	\right]. 
	\label{eq:flow_matching} 
\end{equation} 

The population-optimal velocity field is 
$v^{*}(\bm{z}_{\tau},\tau,c) 
= 
\mathbb{E}[\bm{u}_{\tau}\mid\bm{z}_{\tau},\tau,c]$. 
At inference, the ODE 
$\mathrm{d}\bm{z}_{\tau}/\mathrm{d}\tau 
= 
v_{\theta}(\bm{z}_{\tau},\tau,c)$ 
is integrated from $\tau=1$ to $\tau=0$, and the resulting latent is decoded by the VAE. We adopt Wan2.1-I2V-14B~\citep{wan2025wan}, which combines a causal 3D VAE with a conditional diffusion transformer, as our generative backbone.

\begin{figure}[t]
	\centering
	\vspace{-5pt}
	\includegraphics[width=1.0\linewidth]{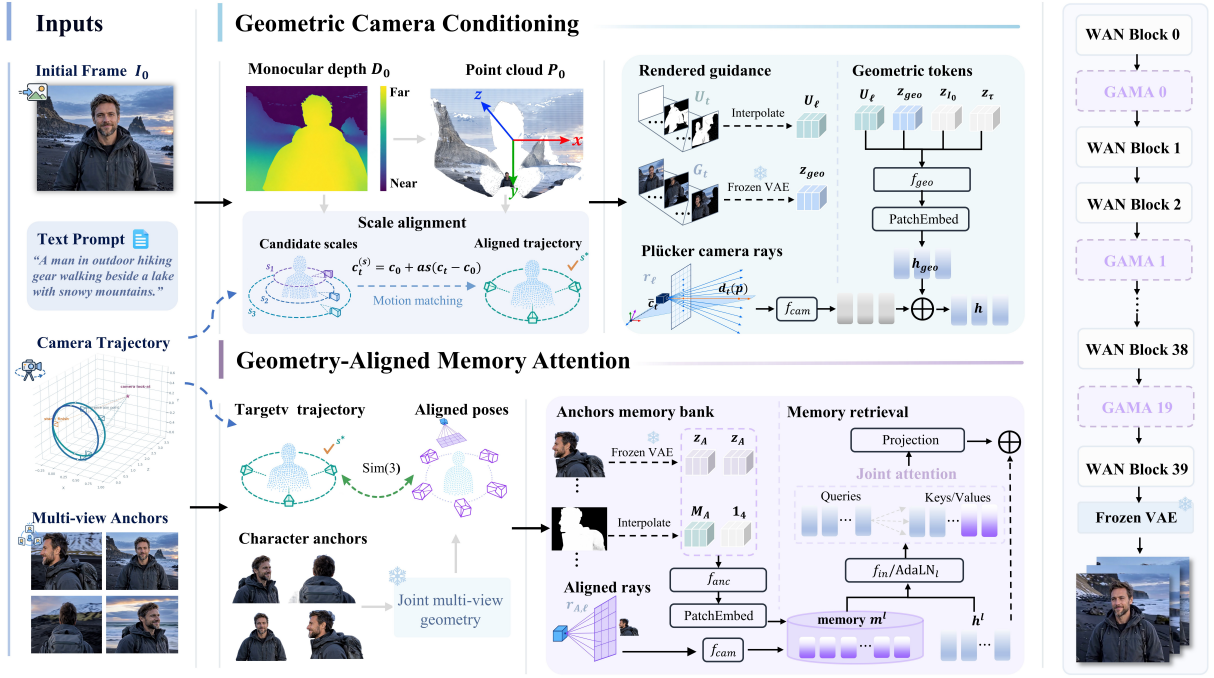}
	\vspace{-12pt}
	\caption{
		Overview of \textit{\method}. The framework combines trajectory-aligned geometric guidance with geometry-aligned multi-view character memory.
	}
	\label{fig:framework}
	\vspace{-5pt}
\end{figure}

\subsection{Problem Formulation}
\label{sec:problem}

We formulate character-centric video generation under
prescribed camera motion as a conditional generation
problem~\citep{he2025cameractrl}.
The inputs comprise an initial character image
$I_0\in\mathbb{R}^{H\times W\times 3}$ captured from an
arbitrary viewpoint, a text prompt $y$, and a target camera
sequence $\mathcal{C}=\{(\bm{K}_t,\bm{T}_t)\}_{t=0}^{N-1}$,
where $\bm{K}_t\in\mathbb{R}^{3\times 3}$ denotes the camera
intrinsics and $\bm{T}_t\in\mathrm{SE}(3)$ the camera-to-world
pose. The goal is to synthesize future frames
$\mathcal{X}=(I_1,\ldots,I_{N-1})$ that follow the prescribed
trajectory while preserving character geometry, appearance,
and identity and satisfying the text prompt.

Large-angle camera motion reveals character surfaces
unobserved in $I_0$, coupling trajectory fidelity with
the completion of missing character
information~\citep{li2025realcam}.
To provide complementary evidence, we additionally condition
generation on $M$ views of the same character,
$\mathcal{A}=\{A_j\}_{j=1}^{M}$, and define the available
observations as $\mathcal{O}=\{I_0\}\cup\mathcal{A}$.
The conditional generation problem is:
\begin{equation}
	\widehat{\mathcal{X}}
	\sim
	p_{\theta}\!\left(
	\mathcal{X}
	\,\middle|\,
	I_0,y,\mathcal{C},\mathcal{A}
	\right),
	\qquad
	\widehat{\mathcal{X}}
	=
	\{\widehat{I}_t\}_{t=1}^{N-1}.
	\label{eq:problem_formulation}
\end{equation}
As illustrated in \redfigref{fig:framework}, we combine
geometric camera conditioning
(\redsecref{sec:geometric_conditioning}) with
geometry-aligned character memory
(\redsecref{sec:gama}) to achieve accurate trajectory
following and consistent character geometry, appearance,
and identity across viewpoints.

\subsection{Geometric Camera Conditioning}
\label{sec:geometric_conditioning}

\paragraph{Depth--trajectory scale alignment.}
We estimate the initial-view depth $D_0$ using a pretrained depth
estimator~\citep{bochkovskiy2025depth} and calibrate the relative scale
between the predicted depth and camera translation using the training
reference video~\citep{cao2025mvgenmaster}.

Let $\bm c_t\in\mathbb R^3$ denote the camera center at frame $t$.
For each candidate scale $s\in\mathcal S$, we rescale the camera
translation relative to $\bm c_0$ while keeping the rotations and
$D_0$ fixed:
\begin{equation}
	\bm c_t^{(s)}
	=
	\bm c_0
	+
	as\bigl(\bm c_t-\bm c_0\bigr),
	\qquad
	s\in\mathcal S .
	\label{eq:camera_scale}
\end{equation}
Here, $a$ is the fixed scale factor provided by the camera metadata,
and $\bm c_t^{(s)}$ denotes the rescaled camera center. For each candidate scale, we back-project the initial view into a
colored point cloud and render it along the corresponding trajectory.
We select the scale whose induced image motion best matches that of
the reference video:
\begin{equation}
	s^{*}
	=
	\arg\min_{s\in\mathcal S}
	\operatorname*{median}_{i\in\mathcal V_s}
	\left|
	\log
	\frac{m_i^{\mathrm{render}}(s)}
	{m_i^{\mathrm{ref}}}
	\right|.
	\label{eq:scale_alignment}
\end{equation}
Here, $m_i^{\mathrm{render}}(s)$ and $m_i^{\mathrm{ref}}$ denote the
motion magnitudes of the $i$-th sampled frame pair in the rendered
and reference sequences, respectively, and $\mathcal V_s$ is the set
of valid frame pairs.
Detailed projection derivations and motion-estimation procedures are
provided in \redappref{app:scale_alignment}.

\paragraph{Projected geometric guidance.}
We back-project the initial image into a colored point cloud
~\citep{ren2025gen3c}.
Let $\bm R_0$ and $\bm c_0$ denote the rotation and camera center
of $\bm T_0$, respectively.
For each pixel $\bm p=[u,v]^\top\in\Omega_0$ with valid depth,
let $\widetilde{\bm p}=[u,v,1]^\top$ denote its homogeneous coordinate.
Its world-space position is:
\begin{equation}
	\bm X(\bm p)
	=
	\bm R_0
	\left(
	D_0(\bm p)\bm K_0^{-1}\widetilde{\bm p}
	\right)
	+
	\bm c_0,
	\qquad
	\mathcal P_0
	=
	\left\{
	\bigl(
	\bm X(\bm p),
	I_0(\bm p)
	\bigr)
	\,\middle|\,
	\bm p\in\Omega_0
	\right\},
	\label{eq:initial_pointcloud}
\end{equation}
where $\Omega_0$ denotes the set of pixels with valid depth. We then render $\mathcal P_0$ along the calibrated trajectory using
point splatting and depth buffering:
\begin{equation}
	\bigl(
	G_t,\,
	U_t
	\bigr)
	=
	\mathcal R
	\left(
	\mathcal P_0,\,
	\bm K_t,\,
	\bm T_t^{(s^*)}
	\right),
	\qquad
	t=0,\ldots,N-1,
	\label{eq:geometric_render}
\end{equation}
where $\bm T_t^{(s^*)}$ retains the rotation of $\bm T_t$ and uses
the calibrated camera center from \redeqref{eq:camera_scale}.
Here, $G_t$ provides RGB guidance aligned with the target viewpoint,
while $U_t\in\{0,1\}^{H\times W}$ indicates pixels not supported by
the projected point cloud and therefore left for the generator to complete.

\paragraph{Geometric and camera conditioning.}
We encode the projected sequence
$\mathcal G=(G_t)_{t=0}^{N-1}$
with the pretrained VAE to obtain $\bm z_{\mathrm{geo}}$
and resize the coverage masks to the latent grid as
$\bm U_\ell$~\citep{cao2025uni3c}.
Before patch embedding, we concatenate these conditions with
the noisy video latent $\bm z_\tau$ and the native
initial-frame condition $\bm z_{I_0}$:
\begin{equation}
	h_{\mathrm{geo}}
	=
	\operatorname{PatchEmbed}\!\left(
	f_{\mathrm{geo}}\!\left(
	[\bm z_\tau,\bm z_{I_0},
	\bm z_{\mathrm{geo}},\bm U_\ell]_{\mathrm{ch}}
	\right)
	\right),
	\label{eq:geometric_fusion}
\end{equation}
where $\bm z_{I_0}$ contains the native image-conditioning
latent and known-frame mask, $[\cdot]_{\mathrm{ch}}$ denotes
channel-wise concatenation, and $f_{\mathrm{geo}}$ is a
$1\times1\times1$ convolution that projects the fused features
to the input dimension of the pretrained patch embedding.

We further encode the target camera geometry using
Pl\"ucker rays~\citep{sitzmann2021light} in a shared normalized
coordinate frame:
\begin{equation}
	\bm r_t(\bm p)
	=
	\bigl[
	\bm d_t(\bm p),
	\bar{\bm c}_t\times\bm d_t(\bm p)
	\bigr],
	\qquad
	h
	=
	h_{\mathrm{geo}}
	+
	f_{\mathrm{cam}}(\bm r_\ell),
	\label{eq:camera_conditioning}
\end{equation}
where $\bar{\bm c}_t$ and $\bm d_t(\bm p)$ denote the camera
center and unit ray direction in the normalized frame,
respectively, and $\bm r_\ell$ denotes the ray grids resized
to the latent resolution.
The projection $f_{\mathrm{cam}}$ maps the ray features to the
same token grid and feature dimension as $h_{\mathrm{geo}}$
for element-wise addition.

\subsection{Geometry-Aligned Memory Attention}
\label{sec:gama}

\paragraph{Anchor--trajectory geometric alignment.}
We extract character foreground masks for the initial frame
$I_0$ and multi-view anchors $\mathcal A=\{A_j\}_{j=1}^{M}$
using a pretrained segmentation model~\citep{carion2026sam},
and suppress backgrounds before geometric estimation.
A pretrained multi-view geometry model~\citep{Wang_2026_CVPR}
then jointly estimates their camera poses, recovering the relative
geometry in a shared coordinate frame.
To align the anchor geometry with the target trajectory, we establish
camera-center correspondences between observations and target views,
with $I_0$ as the shared initial view.

Let $\widehat{\bm R}_j$ and $\widehat{\bm c}_j$ denote the estimated
camera-to-world rotation and camera center of observation $j$,
where $j=0$ corresponds to $I_0$.
For each observation used in alignment, $\kappa(j)$ denotes the
corresponding target frame, with $\kappa(0)=0$.
The remaining correspondences are obtained from source-frame metadata
when available, or by nearest-azimuth matching.
Given the valid correspondence set $\mathcal J$, we estimate a
similarity transform using Umeyama alignment~\citep{umeyama1991least}:
\begin{equation}
	(\rho^*,\bm Q^*,\bm b^*)
	=
	\arg\min_{\rho>0,\,\bm Q\in\mathrm{SO}(3),\,\bm b\in\mathbb R^3}
	\sum_{j\in\mathcal J}
	\left\|
	\rho\bm Q\widehat{\bm c}_j+\bm b
	-\bm c_{\kappa(j)}
	\right\|_2^2.
	\label{eq:anchor_similarity}
\end{equation}

After reconciling camera conventions, the estimated transform maps
the anchor geometry into the target-trajectory coordinate frame:
\begin{equation}
	\bm c_j^{\mathrm{aln}}
	=
	\rho^*\bm Q^*\widehat{\bm c}_j+\bm b^*,
	\qquad
	\bm R_j^{\mathrm{aln}}
	=
	\bm Q^*\widehat{\bm R}_j.
	\label{eq:anchor_alignment}
\end{equation}
We then apply the same camera normalization as for the target trajectory,
so that the aligned anchor geometry and target-view geometry are expressed
in a common normalized frame.

\paragraph{Geometry-aligned memory construction.}
For each observation $A\in\mathcal O$, we suppress its background
using the character mask and encode the masked observation with the
pretrained VAE to obtain $\bm z_A=\mathcal E(A)$.
We further construct a Pl\"ucker ray grid from its aligned camera
geometry and form the anchor representation as
\begin{equation}
	h_A
	=
	\operatorname{PatchEmbed}\!\left(
	f_{\mathrm{anc}}\!\left(
	[\bm z_A,\bm 1_4,\bm z_A,\bm M_A]_{\mathrm{ch}}
	\right)
	\right)
	+
	f_{\mathrm{cam}}(\bm r_{A,\ell}),
	\label{eq:anchor_memory}
\end{equation}
where $\bm 1_4$ is a four-channel all-ones mask,
$\bm M_A$ is the character mask at latent resolution, and
$\bm r_{A,\ell}$ is the resized Pl\"ucker ray grid.
The latent $\bm z_A$ occupies both latent slots of the pretrained
I2V input layout, while $f_{\mathrm{anc}}$ projects the fused features
to the patch-embedding input dimension.
The camera projection $f_{\mathrm{cam}}$ is shared with the target-video
branch in \redsecref{sec:geometric_conditioning}.
Finally, the resulting tokens from all observations are concatenated
to form the geometry-aligned character memory.

\paragraph{Geometry-aware memory retrieval.}
We insert $20$ memory-attention blocks after backbone blocks
$0,2,\ldots,38$. At insertion $l$, let $m^{(l)}$ and $h^{(l)}$
denote the memory and video tokens. Their concatenation is projected
from $5120$ to $1024$ dimensions and timestep-modulated:
\begin{equation}
	x^{(l)}
	=
	f_{\mathrm{in}}\!\left(
	[m^{(l)},h^{(l)}]_{\mathrm{seq}}
	\right),
	\qquad
	\widetilde{x}^{(l)}
	=
	\operatorname{AdaLN}_l\!\left(x^{(l)},\tau\right),
	\label{eq:bank_input}
\end{equation}
where $[\cdot]_{\mathrm{seq}}$ denotes token concatenation. Each block performs joint self-attention with eight heads
($d_h=128$). Queries, keys, and values are projected from
$\widetilde{x}^{(l)}$, with RMS normalization and rotary positional
encoding applied to queries and keys. The video-token output is:
\begin{equation}
	o_{\mathrm{vid}}^{(l)}
	=
	\operatorname{softmax}\!\left(
	\frac{
		Q_{\mathrm{vid}}^{(l)}
		\bigl(
		[K_{\mathrm{mem}}^{(l)},K_{\mathrm{vid}}^{(l)}]_{\mathrm{seq}}
		\bigr)^{\!\top}
	}{
		\sqrt{d_h}
	}
	\right)
	[V_{\mathrm{mem}}^{(l)},V_{\mathrm{vid}}^{(l)}]_{\mathrm{seq}}.
	\label{eq:memory_retrieval}
\end{equation}
Video queries retrieve complementary character evidence from the
geometry-aligned memory while retaining video context, with shared
camera encoding making the retrieval view-aware.

\paragraph{Residual memory injection.}
Let $F_l$ denote the complete memory block, including the
feed-forward refinement and timestep-conditioned modulation.
Its residual update is projected to the backbone dimension and
injected only into the video tokens:
\begin{equation}
	\Delta x^{(l)}
	=
	F_l(x^{(l)},\tau)-x^{(l)},
	\qquad
	h^{(l)}
	\leftarrow
	h^{(l)}
	+
	f_{\mathrm{out}}^{(l)}
	\!\left(
	[\Delta x^{(l)}]_{\mathrm{vid}}
	\right).
	\label{eq:memory_injection}
\end{equation}
Where $[\cdot]_{\mathrm{vid}}$ selects the video-token portion
of the joint sequence.
Each $f_{\mathrm{out}}^{(l)}$ is independently zero-initialized,
preserving the pretrained backbone at the start of adaptation.
The updated tokens then continue through the backbone with
progressively integrated geometry-aligned character evidence.

\subsection{Training Data Construction}

\begin{figure*}[t]
	\centering
	\includegraphics[width=\textwidth]{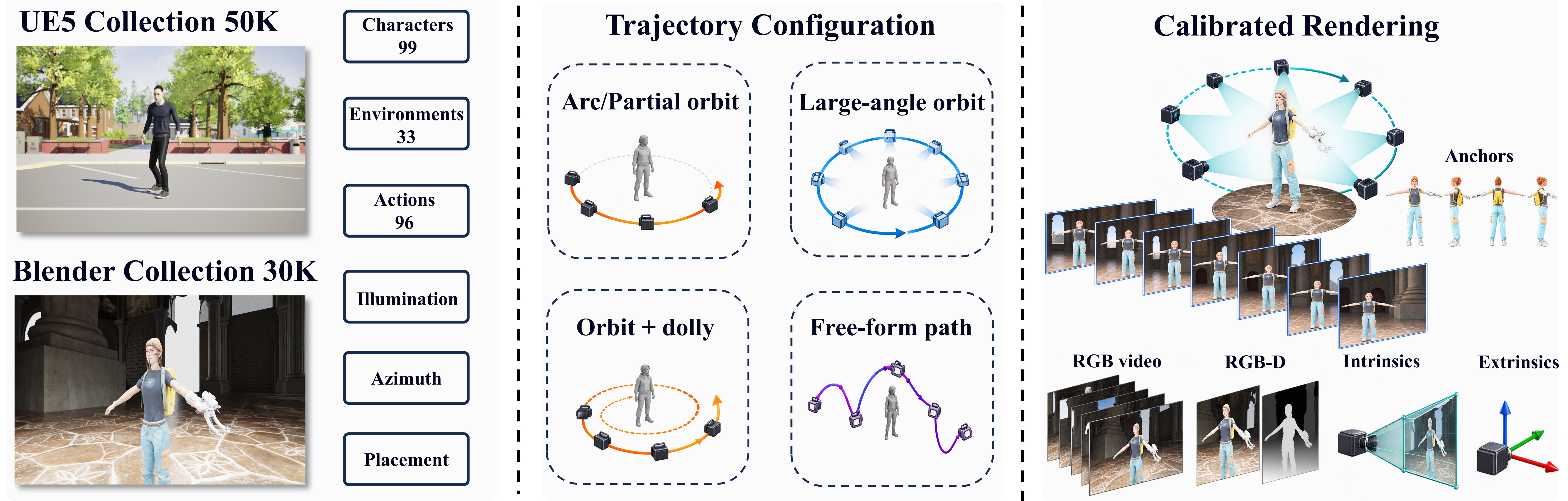}
	\vspace{-16pt}
	\caption{Overview of our training-data construction pipeline.}
	\label{fig:data_construction_overview}
	\vspace{-12pt}
\end{figure*}

\label{sec:training_data}

We construct an $80{,}000$-sample character-centric training corpus, comprising $50{,}000$ UE5 camera-motion sequences and $30{,}000$ Blender-rendered samples with multi-view character anchors, as summarized in \redfigref{fig:data_construction_overview}. Both collections cover $99$ characters and $33$ environments. Using assets from Fab\footnote{\url{https://www.fab.com/listings/2e872a96-686e-4955-8256-82fa19904f92}.}, UE5 combines diverse scenes and character actions with arc, large-angle orbit, orbit-and-dolly, and free-form camera paths. Using assets from Sketchfab\footnote{\url{https://sketchfab.com/}.}, our configurable Blender pipeline automates character placement, camera configuration, preview, and batch rendering across varied initial viewpoints, angular spans, translations, orbit radii, and indoor and outdoor scenes. Each sample provides a reference video, initial RGB-D, camera intrinsics and extrinsics, rendered RGB guidance, coverage masks, and complementary front, back, and side observations for the geometric conditioning and character memory in \redsecref{sec:geometric_conditioning} and \redsecref{sec:gama}. Further construction details and trajectory examples are provided in \redappref{app:data_construction}.

\section{Experiments}
\label{sec:experiments}

\subsection{Experimental Setup}
\label{sec:setup}

\paragraph{Implementation details.}
We train our framework in two stages based on Wan2.1-I2V-14B.
Stage~1 combines $76{,}698$ rendered SpatialVID sequences
~\citep{wang2026spatialvid} with our $50{,}000$ UE5 sequences
for camera-control training.
We update $f_{\mathrm{geo}}$, the pretrained patch embedding,
backbone self-attention layers, and Pl\"ucker camera projection
$f_{\mathrm{cam}}$, while freezing the remaining parameters.
Stage~2 initializes from the Stage~1 checkpoint and trains GAMA
on our $30{,}000$ Blender samples with multi-view character anchors,
optimizing only the memory-attention bank and anchor adapter.

Both stages follow the flow-matching objective in
\redeqref{eq:flow_matching} with their respective conditioning inputs.
We use AdamW with a constant learning rate of $10^{-4}$ and zero
weight decay.
Training uses $81$-frame clips at $832\times480$ resolution.
Stage~1 is trained for $5$ epochs with a batch size of $64$,
and Stage~2 for $8$ epochs with a batch size of $32$.

\paragraph{Evaluation benchmark.}
We evaluate on \textit{\bench{}}, comprising $42$ character-centric scenes:
$17$ synthetic and $25$ real-world cases.
Each case is tested under six camera trajectories spanning large-angle
orbits, translations, and combined motions, with different initial azimuths.
The benchmark evaluates camera-control accuracy, cross-view
character consistency, and visual quality under viewpoint changes.

\paragraph{Compared methods.}
We compare against three groups of methods.
Direct camera-controlled video generation baselines include
CamI2V~\citep{zheng2024cami2v},
Uni3C~\citep{cao2025uni3c},
GEN3C~\citep{ren2025gen3c},
Lyra~2.0~\citep{shen2026lyra}, and
DualCamCtrl~\citep{zhang2025dualcamctrl}.
We further compare with 3D-aware novel-view generation methods,
including ViewCrafter~\citep{yu2024viewcrafter} and
MVGenMaster~\citep{cao2025mvgenmaster},
as well as video re-rendering methods
ReCamMaster~\citep{bai2025recammaster},
DepthDirector~\citep{chen2026beyond}, and
CameraAnything~\citep{li2026cameraanything}.

\paragraph{Evaluation metrics.}
We evaluate camera control using Absolute Trajectory Error (ATE),
Relative Translation Error (RTE), and Relative Rotation Error
(RRE)~\citep{sturm2012evaluating}.
Character consistency is assessed using MEt3R for cross-view
geometric consistency~\citep{asim2025met3r}, ArcFace identity similarity
~\citep{deng2019arcface}, and CLIP and DINO similarities
~\citep{radford2021learning,oquab2023dinov2}.
We report subject consistency (Subj.) and image quality
(IQ) to assess visual fidelity~\citep{huang2024vbench}.

Detailed benchmark configurations, baseline evaluation protocols,
and metric definitions are provided in
\redappref{app:evaluation_details}.

\begin{table*}[t]
	\centering
	\caption{\textbf{Quantitative comparison on our evaluation benchmark.}
		$\downarrow$ and $\uparrow$ denote lower- and higher-is-better metrics, respectively.
		Anchors denotes the use of additional character views.}
	\label{tab:main}
	\scriptsize
	\setlength{\tabcolsep}{6.0pt}
	\renewcommand{\arraystretch}{1.2}
	\begin{tabular}{lcccccccccc}
		\toprule
		\rowcolor{tablehead}
		& &
		\multicolumn{3}{c}{\textbf{Camera Control}}
		& \multicolumn{4}{c}{\textbf{Character Consistency}}
		& \multicolumn{2}{c}{\textbf{Image Quality}} \\
		\cmidrule(lr){3-5}
		\cmidrule(lr){6-9}
		\cmidrule(lr){10-11}
		\rowcolor{tablehead}
		\multirow{-2}{*}{\textbf{Method}}
		& \multirow{-2}{*}{\textbf{Anchors}}
		& \textbf{ATE} $\downarrow$ & \textbf{RTE} $\downarrow$
		& \textbf{RRE} $\downarrow$ & \textbf{MEt3R} $\downarrow$
		& \textbf{Arcf.} $\uparrow$ & \textbf{CLIP} $\uparrow$
		& \textbf{DINO} $\uparrow$ & \textbf{Subj.} $\uparrow$
		& \textbf{IQ} $\uparrow$ \\
		\midrule
		\rowcolor{tablegroup}
		\multicolumn{11}{l}{\textit{Camera-controlled generation}} \\
		CamI2V~\citep{zheng2024cami2v} & \noanchor
		& 0.351 & 0.281 & 11.490 & 0.405 & 0.147 & 0.804 & 0.675 & 0.874 & 0.686 \\
		DualCamCtrl~\citep{zhang2025dualcamctrl} & \noanchor
		& 0.297 & 0.082 & 3.607 & 0.408 & 0.185 & 0.796 & 0.661 & 0.904 & 0.699 \\
		GEN3C~\citep{ren2025gen3c} & \noanchor
		& 0.130 & 0.038 & 1.255 & 0.306 & \underline{0.433}
		& 0.819 & 0.696 & 0.915 & 0.716 \\
		Uni3C~\citep{cao2025uni3c} & \noanchor
		& \underline{0.098} & \underline{0.033} & \underline{1.069}
		& 0.324 & 0.374 & 0.823 & 0.691 & 0.911 & 0.711 \\
		Lyra~2.0~\citep{shen2026lyra} & \noanchor
		& 0.142 & 0.034 & 1.741 & 0.316 & 0.394 & 0.817 & 0.691 & 0.916 & \textbf{0.725} \\
		
		\midrule
		\rowcolor{tablegroup}
		\multicolumn{11}{l}{\textit{Novel-view generation}} \\
		ViewCrafter~\citep{yu2024viewcrafter} & \noanchor
		& 0.353 & 0.203 & 9.198 & 0.395 & 0.223 & 0.810 & 0.683 & 0.888 & 0.715 \\
		ViewCrafter$^{\textcolor{linkred}{\star}}$~\citep{yu2024viewcrafter} & \yesanchor
		& 0.361 & 0.196 & 8.902 & 0.419 & 0.195 & 0.795 & 0.655 & 0.869 & 0.715 \\
		MVGenMaster~\citep{cao2025mvgenmaster} & \noanchor
		& 0.125 & 0.099 & 2.929 & 0.370 & 0.227 & 0.796 & 0.655 & 0.869 & 0.699 \\
		MVGenMaster$^{\textcolor{linkred}{\star}}$~\citep{cao2025mvgenmaster} & \yesanchor
		& 0.218 & 0.152 & 6.938 & \underline{0.301} & 0.299
		& 0.821 & \underline{0.707} & 0.862 & 0.691 \\
		\midrule
		\rowcolor{tablegroup}
		\multicolumn{11}{l}{\textit{Video re-rendering}} \\
		ReCamMaster~\citep{bai2025recammaster} & \noanchor
		& 0.295 & 0.067 & 2.600 & 0.335 & 0.363 & 0.812 & 0.678 & 0.904 & 0.689 \\
		DepthDirector~\citep{chen2026beyond} & \noanchor
		& 0.289 & 0.066 & 1.732 & 0.332 & 0.423
		& \underline{0.826} & 0.686 & \textbf{0.919} & 0.705 \\
		CameraAnything~\citep{li2026cameraanything} & \noanchor
		& 0.244 & 0.078 & 4.879 & 0.341 & 0.358 & 0.807 & 0.648 & 0.898 & 0.700 \\
		\midrule
		\rowcolor{tableours}
		\textit{\method{}} (Ours) & \yesanchor
		& \textbf{0.065} & \textbf{0.020} & \textbf{0.896} & \textbf{0.228}
		& \textbf{0.555} & \textbf{0.830} & \textbf{0.719}
		& \underline{0.917} & \underline{0.719} \\
		\bottomrule
	\end{tabular}
	\vspace{-10pt}
\end{table*}

\begin{figure}[t]
	\centering
	\vspace{-5pt}
	\includegraphics[width=1.0\linewidth]{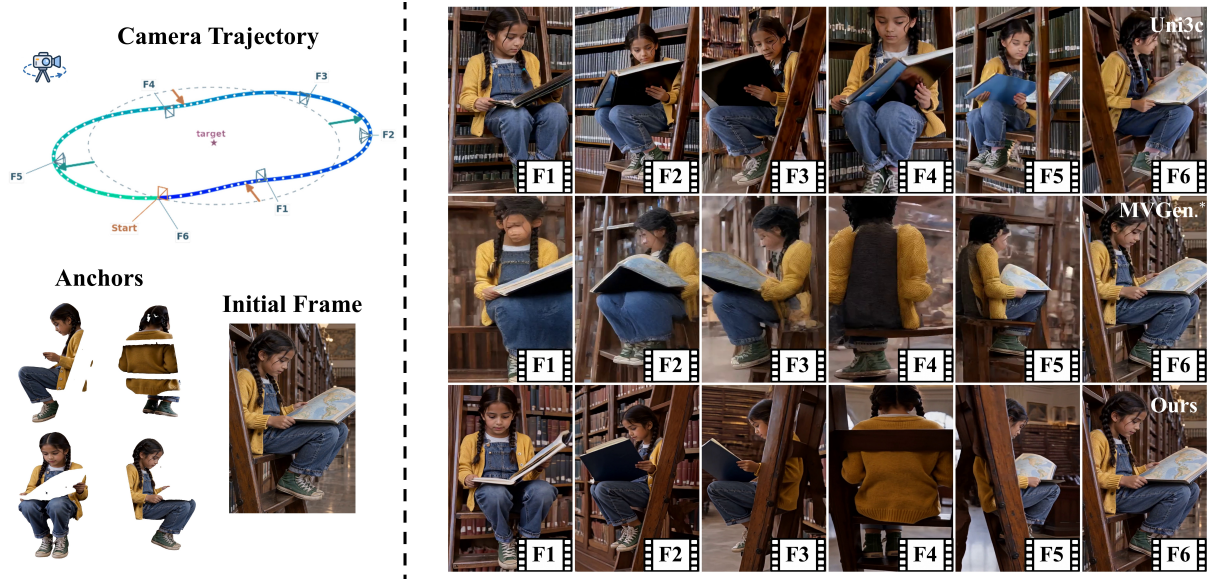}
	\vspace{-12pt}
	\caption{
		Qualitative comparison on large-angle camera-controlled video generation.
	}
	\vspace{-8pt}
	\label{fig4}
\end{figure}

\subsection{Comparison with State-of-the-Art Methods}
\label{sec:quant}

\paragraph{Quantitative comparison.}
As shown in \redtabref{tab:main}, \textit{\method{}} achieves superior camera
control and character consistency while maintaining competitive visual
quality.
The gains span trajectory accuracy, cross-view geometry, and identity
and appearance preservation.
Additional reference views yield limited or inconsistent improvements
for ViewCrafter and MVGenMaster, reflecting the difficulty of reliably
aligning sparse cross-view geometry under large viewpoint changes.
In contrast, our framework combines trajectory-aligned geometric guidance
with geometry-aligned character memory, enabling more reliable large-angle
camera control and character preservation.

\paragraph{Qualitative comparison.}

As shown in \redfigref{fig4}, we evaluate a challenging trajectory combining a large-angle orbit with dolly motion. We compare against Uni3C and MVGenMaster. Uni3C exhibits severe character distortion under large viewpoint changes, particularly in facial regions. MVGenMaster uses the initial frame together with multi-view references containing both character and scene content, yet still loses fine character details and appearance cues. In contrast, our approach follows the prescribed camera motion more reliably and better preserves character identity, 3D structure, and appearance throughout the trajectory.

\subsection{Ablations}
\label{sec:ablation}

\begin{table*}[t]
	\centering
	\caption{\textbf{Ablation study of the proposed components.}
		We remove the entire GAMA branch, anchor geometry alignment,
		or anchor foreground masking while keeping the remaining settings unchanged.}
	\label{tab:module_ablation}
	\scriptsize
	\setlength{\tabcolsep}{6.0pt}
	\renewcommand{\arraystretch}{1.2}
	\begin{tabular}{lccccccccc}
		\toprule
		\rowcolor{tablehead}
		&
		\multicolumn{3}{c}{\textbf{Camera Control}}
		& \multicolumn{4}{c}{\textbf{Character Consistency}}
		& \multicolumn{2}{c}{\textbf{Visual Quality}} \\
		\cmidrule(lr){2-4}
		\cmidrule(lr){5-8}
		\cmidrule(lr){9-10}
		\rowcolor{tablehead}
		\multirow{-2}{*}{\textbf{Variant}}
		& \textbf{ATE} $\downarrow$
		& \textbf{RTE} $\downarrow$
		& \textbf{RRE} $\downarrow$
		& \textbf{MEt3R} $\downarrow$
		& \textbf{Arcf.} $\uparrow$
		& \textbf{CLIP} $\uparrow$
		& \textbf{DINO} $\uparrow$
		& \textbf{Subj.} $\uparrow$
		& \textbf{IQ} $\uparrow$ \\
		\midrule
		w/o GAMA
		& 0.100 & 0.031 & 1.149
		& 0.333 & 0.370 & 0.813 & 0.668
		& 0.899 & 0.699 \\
		w/o Anchor Geometry Alignment
		& 0.071 & 0.027 & 1.374
		& 0.239 & 0.547 & 0.826 & 0.708
		& 0.909 & 0.716 \\
		w/o Anchor Masking
		& 0.078 & 0.025 & 1.110
		& 0.247 & 0.546 & 0.827 & 0.710
		& 0.908 & 0.709 \\
		\midrule
		\rowcolor{tableours}
		\textit{\method{}} {(Ours)}
		& \textbf{0.065} & \textbf{0.020} & \textbf{0.896}
		& \textbf{0.228} & \textbf{0.555}
		& \textbf{0.830} & \textbf{0.719}
		& \textbf{0.917} & \textbf{0.719} \\
		\bottomrule
	\end{tabular}
	\vspace{-5pt}
\end{table*}

% Preamble: \usepackage{wrapfig}

\begin{wraptable}{r}{0.50\linewidth}
	\centering
	\vspace{-15pt}
	\caption{\textbf{First-stage camera-control ablation.}
	}
	\label{tab:stage1_ablation}
	\scriptsize
	\setlength{\tabcolsep}{6.0pt}
	\renewcommand{\arraystretch}{1.15}
	\begin{tabularx}{\linewidth}{
			>{\raggedright\arraybackslash}X ccc
		}
		\toprule
		\rowcolor{tablehead}
		& \multicolumn{3}{c}{\textbf{Camera Control}} \\
		\cmidrule(lr){2-4}
		\rowcolor{tablehead}
		\multirow{-2}{*}{\textbf{Variant}}
		& \textbf{ATE} $\downarrow$
		& \textbf{RTE} $\downarrow$
		& \textbf{RRE} $\downarrow$ \\
		\midrule
		
		w/o UE5 data
		& 0.359 & 0.156 & 4.295 \\
		
		w/o Pl\"ucker conditioning
		& 0.275 & 0.056 & 2.113 \\
		
		\midrule
		\rowcolor{tableours}
		\textbf{Full Stage~1}
		& \textbf{0.100}
		& \textbf{0.031}
		& \textbf{1.149} \\
		\bottomrule
	\end{tabularx}
	\vspace{-5pt}
\end{wraptable}

\paragraph{Effect of individual components.} As shown in \redtabref{tab:module_ablation}, the full configuration achieves the best performance across all metrics. Removing GAMA substantially degrades both camera control and character consistency, confirming the importance of multi-view character memory. Without anchor geometry alignment, rotation error increases markedly while identity similarity remains relatively high, highlighting the importance of relating anchor observations to target viewpoints. Removing foreground masking also degrades camera control, character consistency, and visual quality, demonstrating the benefit of isolating character information from scene backgrounds.

\paragraph{Effect of UE5 data and camera-ray conditioning.}
We further examine the contributions of UE5 training data
and Pl\"ucker ray conditioning to camera control.
These ablations are conducted on the first-stage model
before GAMA adaptation.
As shown in \redtabref{tab:stage1_ablation}, excluding
the UE5 collection increases all three camera errors,
demonstrating the benefit of large-angle training sequences.
Removing Pl\"ucker conditioning also degrades camera accuracy,
showing that explicit camera-ray information effectively complements
the projected geometric guidance under large-angle camera motion.

\begin{wrapfigure}{r}{0.52\linewidth}
	\centering
	\includegraphics[width=0.95\linewidth]{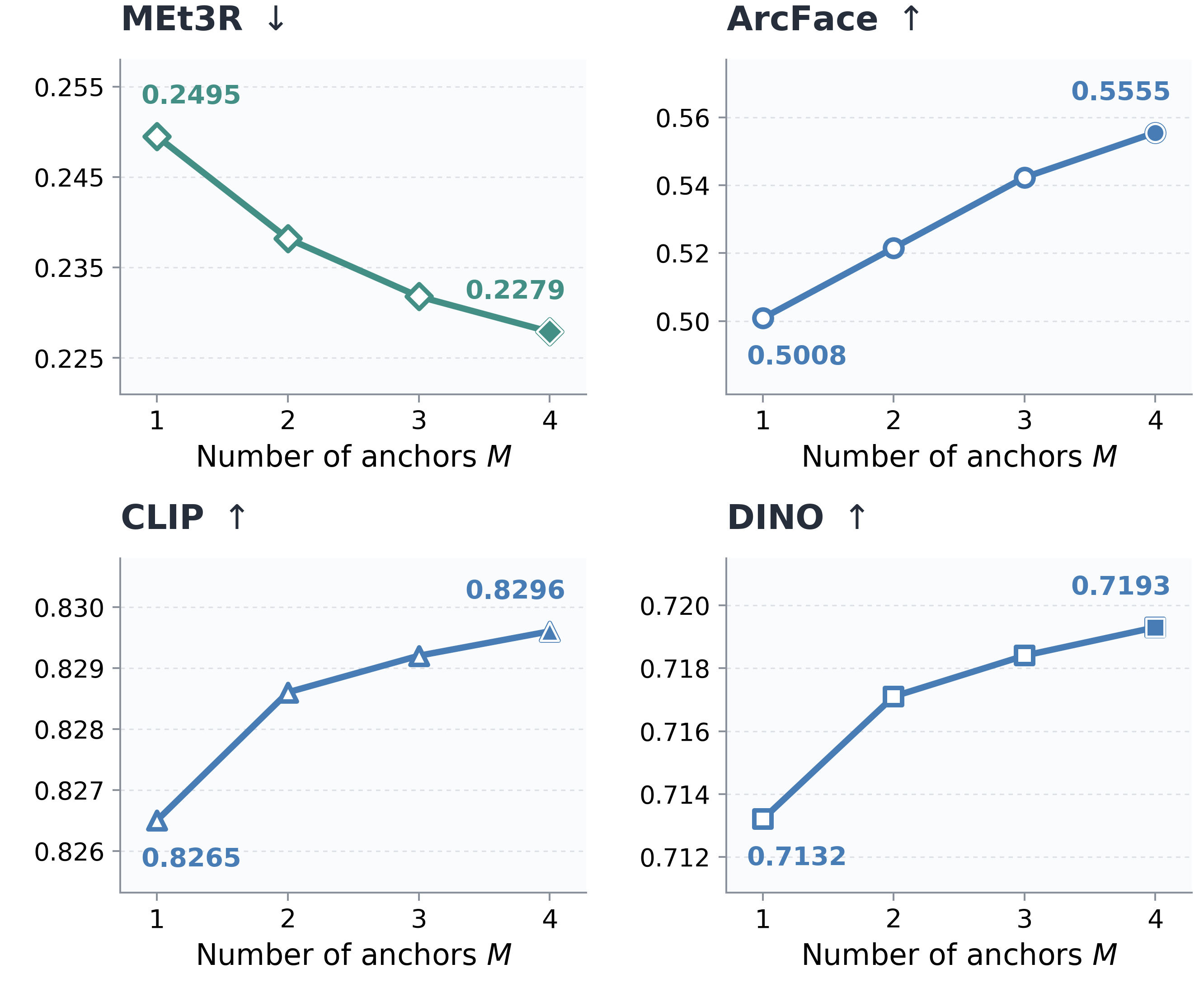}
	\vspace{-5pt}
	\caption{Effect of the number of character anchors on character consistency.}
	\label{fig:anchor_ablation}
	\vspace{-40pt}
\end{wrapfigure}

\paragraph{Effect of anchor count.}
As shown in \redfigref{fig:anchor_ablation}, increasing the number of
character anchors from one to four improves cross-view geometry,
identity preservation, and appearance similarity under large viewpoint
changes.
This indicates that complementary observations help recover character
information missing from the initial view and provide richer cross-view
evidence.
The smaller gain from three to four anchors suggests diminishing returns
as additional views contribute less new information.

Additional discussions, supplementary experiments, and more qualitative
visualizations are provided in \redappref{app:additional_experiments}.

\section{Conclusion}
\label{sec:conclusion}

We presented \textit{\bench{}}, a character-centric dataset and benchmark for
large-angle camera-controlled video generation, together with \textit{\method{}},
a framework that jointly improves trajectory accuracy and character
preservation under substantial viewpoint changes.
The key idea is to combine scale-aligned geometric guidance with
Geometry-Aligned Memory Attention (GAMA), which aligns multi-view
character observations with the target camera geometry and retrieves
view-relevant character features during generation.
Experiments on both synthetic and real-world scenes demonstrate more
accurate camera control and stronger preservation of character geometry,
appearance, and identity, while maintaining competitive visual quality.
Ablation studies further confirm the importance of large-angle training
data, explicit camera-ray conditioning, geometric anchor alignment, and
multi-view character memory.
Future work will explore more flexible anchor acquisition, more complex
dynamic scenes, and longer or interactive camera trajectories.

\bibliography{ref}

\subsection*{AI Use Statement}

Generative AI tools were used to assist with language polishing,
translation, and the revision of scientific writing, including improving
clarity, grammar, and presentation.
All AI-assisted content was reviewed and revised by the authors.
Technical claims, mathematical formulations, experimental settings,
citations, and implementation details were verified against the
corresponding source materials and our implementation.
Generative AI was not used to autonomously conduct experiments,
determine experimental results, or make scientific conclusions.
The authors take full responsibility for the final content of this work,
including all text, claims, code, and artifacts produced with the aid of
generative AI.

\subsection*{Ethics Statement}

This work studies controllable character video generation and therefore
has potential applications in visual-content creation as well as potential
risks associated with synthetic human imagery.
Our training benchmark is constructed primarily from controlled
simulations using licensed or publicly available 3D character and
environment assets.
Any released real-world evaluation data will exclude personally
identifiable private individuals and will follow the applicable data
licenses and usage terms.
Although our method is designed for research on camera control and
character consistency, character-generation technology could be misused
to create misleading or impersonating content.
We therefore discourage the synthesis of deceptive depictions of real
individuals without their knowledge or consent and will provide
provenance and documentation for released benchmark assets and generated
examples where applicable.

\subsection*{Reproducibility Statement}

We provide the architectural design, conditioning formulation, training
strategy, optimization settings, and evaluation protocol in the main
paper and appendices.
The supplementary material further documents the construction of the
UE5 and Blender datasets, camera and depth alignment procedures,
baseline input adaptations, evaluation metrics, and additional ablations.
The benchmark generator exports camera intrinsics, camera-to-world poses,
depth maps, multi-view character anchors, reference videos, point-cloud
renders, and coverage masks.
Upon publication, we plan to release the benchmark generation and
evaluation code, pose-alignment and identity-evaluation procedures,
trajectory configurations, training and inference configurations, and
the settings required to reproduce results with the released checkpoint.

\clearpage
\appendix

% Appendix numbering: A.1, A.2, B.1, ...
\counterwithin{figure}{section}
\counterwithin{table}{section}
\numberwithin{equation}{section}

\section{Depth--Trajectory Scale Calibration}
\label{app:scale_alignment}

This section details the reprojection geometry and
motion-estimation procedure for the scale calibration
in \redsecref{sec:geometric_conditioning}.
Predicted depth and camera translation may follow different
scale conventions. We calibrate their relative scale by
matching the motion induced by point-cloud rendering
to that of the training reference video.

\paragraph{Reprojection geometry.}
Let $D_0$ denote the predicted depth of the initial frame, and let
$\bm R_t$ and $\bm c_t$ denote the camera-to-world rotation and camera
center at frame $t$, respectively.
For an initial-view pixel $\bm p=[u,v]^\top$, we define its homogeneous
coordinate as $\widetilde{\bm p}=[u,v,1]^\top$.
The corresponding 3D point in the initial camera frame is
\begin{equation}
	\bm q_0(\bm p)
	=
	D_0(\bm p)\bm K_0^{-1}\widetilde{\bm p}.
	\label{eq:app_backprojection}
\end{equation}

Following \redeqref{eq:camera_scale}, the camera translation is rescaled
while the rotations and predicted depth remain fixed.
The point is transformed from the initial camera frame to the target
camera frame as
\begin{align}
	\bm q_t^{(s)}(\bm p)
	&=
	\bm R_t^\top
	\left(
	\bm R_0\bm q_0(\bm p)
	+\bm c_0-\bm c_t^{(s)}
	\right)
	\nonumber\\
	&=
	\bm R_t^\top\bm R_0\bm q_0(\bm p)
	+
	as\bm R_t^\top(\bm c_0-\bm c_t).
	\label{eq:app_target_point}
\end{align}
Using perspective projection
$\pi([x,y,z]^\top)=[x/z,y/z]^\top$, the corresponding target pixel is
\begin{equation}
	\bm p_t^{(s)}
	=
	\pi\!\left(
	\bm K_t\bm q_t^{(s)}(\bm p)
	\right).
	\label{eq:app_pixel_projection}
\end{equation}

Substituting \redeqref{eq:app_backprojection} and exploiting the
scale invariance of perspective projection gives
\begin{equation}
	\bm p_t^{(s)}
	=
	\pi\!\left(
	\bm K_t
	\left[
	\bm R_t^\top\bm R_0
	\bm K_0^{-1}\widetilde{\bm p}
	+
	\frac{as}{D_0(\bm p)}
	\bm R_t^\top(\bm c_0-\bm c_t)
	\right]
	\right).
	\label{eq:app_projection}
\end{equation}
This expression makes the relevant scale ambiguity explicit:
for fixed rotations and intrinsics, the translational component of
image motion depends on camera displacement relative to scene depth.
Varying $s$ therefore changes the induced parallax without altering
the predicted depth structure or camera rotations.

\paragraph{Motion estimation.}
We back-project the initial frame into a colored point cloud.
For each candidate scale $s\in\mathcal S$, we render this
point cloud along the rescaled trajectory, obtaining
$\mathcal G^{(s)}=(G_t^{(s)})_{t=0}^{N-1}$.
Point splatting with depth buffering handles visibility.
We then estimate image motion independently from the rendered sequence
$\mathcal G^{(s)}$ and the training reference sequence
$\mathcal I^{\mathrm{ref}}=(I_t^{\mathrm{ref}})_{t=0}^{N-1}$
at the same preview resolution.

We sample frame pairs $(t_i,t_i+\Delta)$, where
$t_i=i\Delta$, $t_i+\Delta<N$, and $\Delta=4$ by default.
For each pair, up to $1200$ corners are detected in the first frame
and tracked to the second using pyramidal Lucas--Kanade optical
flow~\citep{lucas1981iterative}.
Pairs with fewer than $16$ detected corners or valid tracks are discarded.
To suppress unreliable correspondences, we fit a partial affine model
with RANSAC~\citep{fischler1981random} using a $3$-pixel reprojection
threshold.
RANSAC inliers are retained when at least $12$ are available;
otherwise, all valid tracks are used.

For a sequence $V$, let $\bm p_{i,k}(V)$ and
$\bm p'_{i,k}(V)$ denote the coordinates of retained track $k$
in the two frames of pair $i$, and let $\mathcal J_i(V)$ denote
the corresponding track-index set.
We define the motion magnitude of the pair as
\begin{equation}
	m_i(V)
	=
	\operatorname*{median}_{k\in\mathcal J_i(V)}
	\left\|
	\bm p'_{i,k}(V)-\bm p_{i,k}(V)
	\right\|_2.
	\label{eq:app_motion}
\end{equation}
Applying the same estimator to both sequences yields
$m_i^{\mathrm{render}}(s)=m_i(\mathcal G^{(s)})$ and
$m_i^{\mathrm{ref}}=m_i(\mathcal I^{\mathrm{ref}})$.

\paragraph{Scale selection.}
For each candidate $s$, we retain frame pairs whose motion estimates
are finite and larger than $\epsilon=10^{-3}$ pixels:
\begin{equation}
	\mathcal V_s
	=
	\left\{
	i \;\middle|\;
	m_i^{\mathrm{render}}(s)\in(\epsilon,\infty),
	\;
	m_i^{\mathrm{ref}}\in(\epsilon,\infty)
	\right\}.
	\label{eq:app_valid_pairs}
\end{equation}
We evaluate \redeqref{eq:scale_alignment} over the candidate set
\begin{equation}
	\mathcal S
	=
	\{0.05,0.075,0.10,0.15,0.20,0.25,0.33,0.40,0.50\},
	\label{eq:app_scale_candidates}
\end{equation}
and assign an infinite score to candidates with fewer than four valid
frame pairs.
The absolute log ratio measures multiplicative motion mismatch
symmetrically, while the median reduces sensitivity to atypical frame
pairs. 

The selected scale $s^*$ is substituted into
\redeqref{eq:camera_scale} to construct the geometric
conditions for the training clip.
During training, inconsistent depth and translation scales
cause the rendered parallax to deviate from that of the
ground-truth video, weakening the alignment between
geometric guidance and supervision.
Calibration reduces this discrepancy by matching their
image-motion magnitudes.
At inference, camera translation is interpreted relative
to the estimated scene depth, so the initial-view point
cloud and requested trajectory jointly define the target
parallax.
No reference-based scale calibration is required, as there
is no independently observed video motion to match.

\section{Training Data Construction}
\label{app:data_construction}

\begin{figure*}[h]
	\centering
	\includegraphics[width=\textwidth]{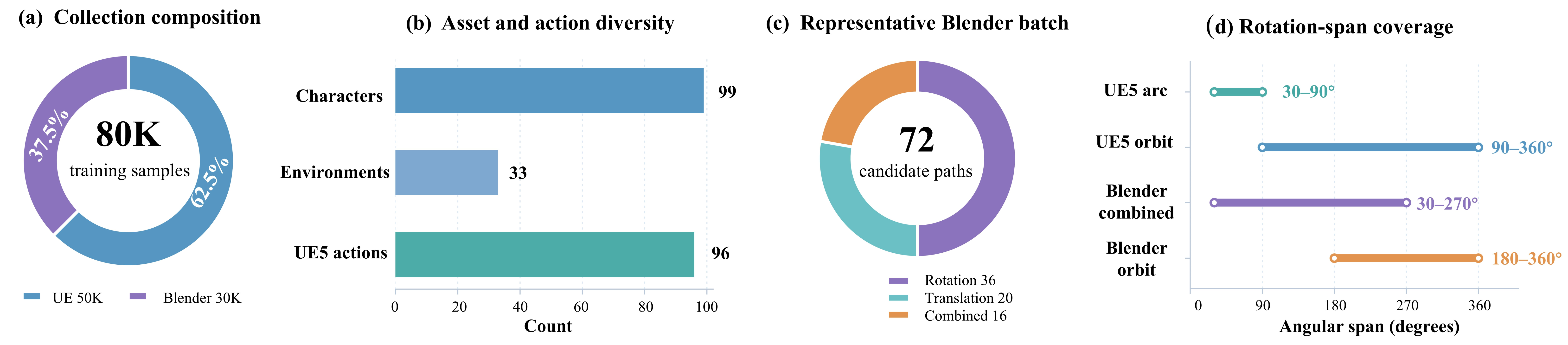}
	\caption{
		\textbf{Training-data composition and camera configurations.}
		(a) Sizes and training roles of the UE5 and Blender collections.
		(b) Character and environment counts shared by both collections,
		together with the number of UE5 character actions.
		(c) A representative Blender batch containing $36$ rotational,
		$20$ translational, and $16$ combined trajectories.
		(d) Angular-span configurations of the camera trajectories.
	}
	\label{fig:data_statistics}
\end{figure*}

We construct two complementary training collections: UE5 sequences
for large-angle camera-control learning and Blender samples with
multi-view anchors for character-memory learning.
\redtabref{tab:training_data_configuration} summarizes their training
roles and camera configurations, while
\redfigref{fig:data_statistics} shows their scale, asset statistics,
and representative trajectory settings.

\subsection{UE5 Character-Centric Camera-Motion Data}

\paragraph{Scenes and character actions.}
We render $50{,}000$ character-centric sequences using
$99$ character models, $33$ environments, and $96$ character actions.
The environments span indoor and outdoor settings, including city
streets, shopping malls, cafes, offices, and rural scenes, while the
actions include motions such as waving, dancing, and cheering.
Each sequence contains a single character, with the character, action,
environment, and camera configuration varied across the collection.
For each character--environment configuration, we additionally render
front, back, left, and right views as complementary observations.

\paragraph{Camera trajectories.}
We design four trajectory families: arcs spanning
$30^\circ$--$90^\circ$, large-angle orbits spanning
$90^\circ$--$360^\circ$, orbit-and-dolly paths combining
$90^\circ$--$360^\circ$ sweeps with camera-distance variation,
and free-form trajectories generated by B-spline interpolation
between multiple keyframes.
Both constant- and variable-speed camera motions are included.
The configurations of the UE5 and Blender collections are summarized
in \redtabref{tab:training_data_configuration}, and representative
angular spans are shown in \redfigref{fig:data_statistics}(d).

\begin{table}[h]
	\centering
	\caption{Comparison of the UE5 and Blender training collections.}
	\label{tab:training_data_configuration}
	\small
	\setlength{\tabcolsep}{7pt}
	\renewcommand{\arraystretch}{1.22}
	\begingroup
	\renewcommand{\tabularxcolumn}[1]{m{#1}}
	\begin{tabularx}{\linewidth}{
			@{} >{\centering\arraybackslash}m{0.20\linewidth}
			>{\centering\arraybackslash}X
			>{\centering\arraybackslash}X @{}
		}
		\toprule
		\textbf{Configuration}
		& \textbf{UE5}
		& \textbf{Blender} \\
		\midrule
		
		Training samples
		& $50{,}000$
		& $30{,}000$ \\
		
		Training stage
		& Large-angle camera control
		& Multi-view memory injection \\
		
		Anchor conditioning
		& Not used
		& Front, back, left, and right views \\
		
		\midrule
		
		Camera paths
		& Arcs, orbits, orbit-and-dolly,
		and free-form paths
		& Orbits, translations,
		and combined rotation--translation paths \\
		
		Motion variation
		& Angular span, camera distance,
		and speed
		& Initial azimuth, angular span,
		translation, and orbit radius \\
		
		\bottomrule
	\end{tabularx}
	\endgroup
	
	\vspace{4pt}
	\parbox{\linewidth}{\footnotesize
		Both collections use $99$ character models and
		$33$ indoor and outdoor environments.}
\end{table}

\subsection{Blender Data with Multi-View Character Anchors}

\paragraph{Asset preparation and scene placement.}
We generate $30{,}000$ samples using an automated Blender pipeline
with Cycles rendering.
The assets include indoor and outdoor environments as well as static,
rigged, and animated characters.
The pipeline computes scene bounds, rescales characters, and evaluates
candidate placements based on ground height, visibility, and clearance.
Character placement and camera framing are adjusted to maintain ground
contact and keep the subject within view.
Orbit centers are derived from the placed character geometry, while
camera radius and focal length are selected according to character size
and scene constraints.

\paragraph{Trajectory generation.}
Camera trajectories are parameterized by initial azimuth, angular span,
orbit radius, and translation direction and distance.
The pipeline supports $180^\circ$--$360^\circ$ orbits, partial orbits
from different initial viewpoints, pure translations, combined
rotation--translation paths, and smooth radius variation.
A representative batch contains $36$ rotational, $20$ translational,
and $16$ combined paths, yielding $72$ candidate trajectories per
character--environment pair, as illustrated in
\redfigref{fig:data_statistics}(c).

\paragraph{Multi-view anchors and geometric outputs.}
We render front, back, left, and right observations of each character
as multi-view anchors, together with the reference video, initial RGB
image, initial-view depth, and per-frame camera intrinsics and extrinsics.
The initial RGB-D observation is back-projected into a colored point
cloud and rendered along the recorded trajectory to produce projected
RGB guidance and masks indicating regions unsupported by the initial view.
The initial image and multi-view anchors are further processed by the
camera-estimation and geometric-alignment procedure in
\redsecref{sec:gama} to construct the geometry-aligned character memory.

\paragraph{Preview and batch rendering.}
A configurable web interface supports asset selection, camera adjustment,
four-view preview, and batch rendering.
The previews enable inspection of framing, occlusion, and ground contact
before full-sequence rendering.
Persistent Blender workers reuse loaded assets across tasks, while asset
identifiers, camera parameters, and rendering settings are recorded for
reproducibility.

\section{Benchmark and Evaluation Details}
\label{app:evaluation_details}

This section provides additional details of the evaluation benchmark,
baseline input configurations, and metric protocol used in
\redsecref{sec:experiments}.

\subsection{Benchmark Construction}

\paragraph{Evaluation scenes.}
The evaluation set contains $42$ character-centric scenes, including
$17$ synthetic scenes rendered in UE5 and $25$ collected from
real-world imagery.
The two subsets provide complementary evaluation settings: synthetic
cases offer controlled character and scene configurations, while
real-world cases assess generalization to natural images and diverse
visual content.
All evaluation cases are disjoint from the training collections.

\paragraph{Camera trajectories.}
Each synthetic or real-world scene is evaluated under six prescribed
camera trajectories covering large-angle orbits, pure translations,
and combined rotation--translation motions.
These trajectories span substantial viewpoint changes and diverse
initial azimuths, varying the character content observed in the initial
view and enabling joint evaluation of trajectory adherence and character
consistency as previously unseen regions become visible.

\paragraph{Evaluation inputs.}
Each evaluation case provides an initial character image and a prescribed
camera trajectory.
For methods that support or require multiple reference views, we
additionally provide multi-view character observations.
Our method takes the initial image together with the corresponding
multi-view character anchors.
All compared methods follow the same target trajectories, with camera
parameters converted to their native representations when necessary.
Generation resolution and sequence length follow the native settings
of each method.

\subsection{Baseline Evaluation Protocol}

The compared methods span three generation paradigms.
We adapt the benchmark inputs to their native formulations while
preserving the prescribed target camera motion.

\paragraph{Direct camera-controlled generation.}
Camera-controlled image-to-video methods are provided with the initial
image and prescribed camera trajectory following their native input
formats.

\paragraph{Novel-view generation.}
For novel-view synthesis methods, we evaluate both their native single-image setting and, where supported, a multi-image setting with character anchors. For the latter, we follow each method's native multi-view formulation and provide the initial image together with the character anchors as reference views. Target views are then synthesized along the prescribed camera trajectory. This setting allows us to separately examine the effect of additional reference observations when comparing with our multi-view character conditioning.

\paragraph{Video re-rendering.}
Video re-rendering methods take a source video rather than a single initial image as input. We therefore construct a static source video by repeating the initial image along the temporal dimension and apply the same prescribed camera trajectory. This setup isolates camera-induced changes from source-video motion and provides a common static-input setting for comparison with image-conditioned methods.

\paragraph{Generation settings.}
Each method is evaluated using the generation resolution and sequence
length supported by its native implementation.
All generated outputs are then evaluated using the same quantitative
evaluation pipeline.

\subsection{Evaluation Metrics}

We evaluate each method along three complementary dimensions:
camera-control accuracy, character consistency, and visual quality.

\paragraph{Camera-control accuracy.}
We recover the camera trajectory of each generated video using
Depth Anything~3~\citep{lin2025depth}. To remove the global coordinate-frame and scale ambiguity of monocular
pose estimation, we align each predicted trajectory to its target
trajectory using a Sim(3) Umeyama transformation.
Following standard trajectory-evaluation practice
~\citep{sturm2012evaluating}, we compute Absolute Trajectory Error
(ATE), Relative Translation Error (RTE), and Relative Rotation Error
(RRE).
ATE measures the RMSE between aligned and target camera centers, while
RTE and RRE measure the translational and rotational errors of
consecutive relative poses, respectively.
Lower values indicate better camera adherence.

\paragraph{Character consistency.}
We use MEt3R~\citep{asim2025met3r} to measure cross-view geometric
consistency using all generated frames together with the four
observation anchors.
We further evaluate identity and appearance consistency using ArcFace
~\citep{deng2019arcface}, CLIP~\citep{radford2021learning}, and
DINO~\citep{oquab2023dinov2} similarities.
To account for large viewpoint changes, for each generated frame, we
select the observation anchor with the closest camera pose and compute
its similarity to that anchor.
The resulting similarities are averaged over all frames and videos.
Higher ArcFace, CLIP, and DINO similarities indicate stronger
consistency, while lower MEt3R indicates better geometric consistency.

\paragraph{Visual quality.}
We additionally report subject consistency (Subj.) and image quality
(IQ) following their default evaluation settings
~\citep{huang2024vbench} to assess the overall visual fidelity of
generated videos.

\section{Additional Analyses and Visualizations}
\label{app:additional_experiments}

This section provides additional experimental analyses
and qualitative visualizations to further evaluate the proposed framework.

\begin{figure}[t]
	\centering
	\vspace{-5pt}
	\includegraphics[width=0.7\linewidth]{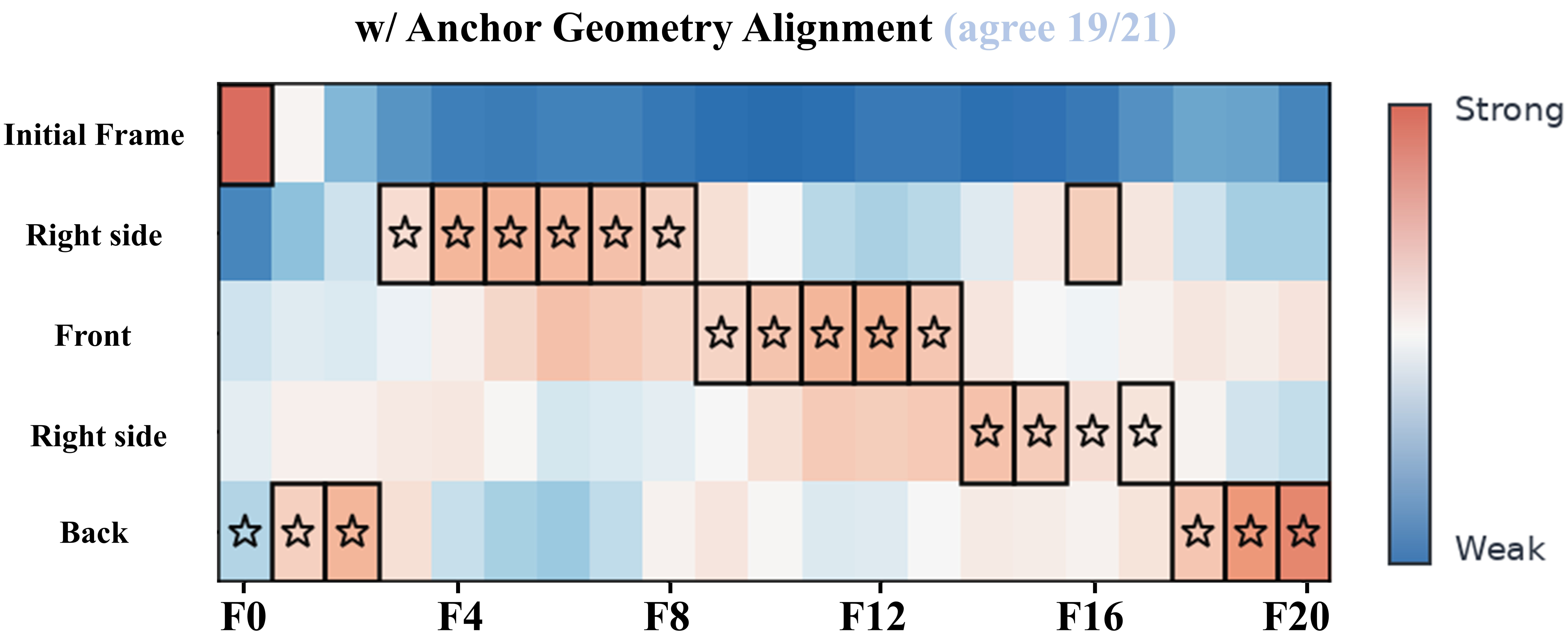}
	\vspace{-12pt}
	\caption{
		\textbf{View-aware anchor routing under a $360^\circ$ camera orbit.}
		Attention responses are visualized for the $21$ temporal latent positions
		corresponding to an $81$-frame generated video.
		Stars denote the geometrically expected observations with the closest
		camera poses to the target viewpoints, while outlined cells indicate the
		observations receiving the strongest attention responses.
		The predicted routing agrees with the expected routing for $19/21$
		latent positions.
	}
	\label{fig:anchor_routing_map}
	\vspace{-5pt}
\end{figure}

\begin{figure}[t]
	\centering
	\vspace{-0pt}
	\includegraphics[width=0.6\linewidth]{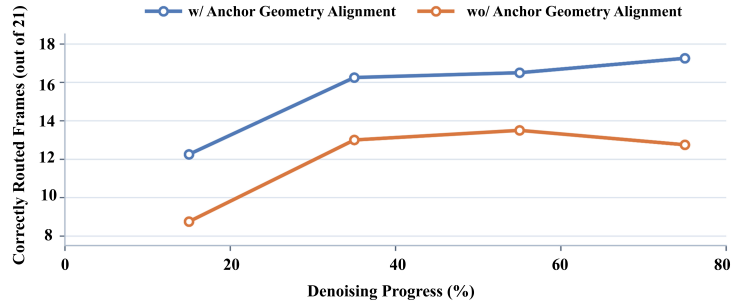}
	\vspace{-4pt}
	\caption{
		\textbf{Effect of Anchor Geometry Alignment on attention routing.}
		Routing accuracy during training with and without anchor geometry alignment.
		Geometry alignment enables faster learning of view-aware anchor retrieval
		and achieves higher routing accuracy after convergence.
	}
	\label{fig:anchor_routing_training}
	\vspace{-5pt}
\end{figure}

\subsection{Analysis of Geometry-Aligned Anchor Routing}
\label{app:anchor_routing}

\paragraph{View-aware anchor routing.}
We analyze how the generated video features retrieve character information
from different observations under a $360^\circ$ camera orbit.
As shown in \redfigref{fig:anchor_routing_map}, the $81$ generated frames
correspond to $21$ temporal latent positions, each attending to the initial
frame and multi-view character anchors.
For each latent position, the star marks the expected anchor whose camera
pose is closest to the target viewpoint, while the outlined cell denotes
the observation receiving the strongest attention response.
We observe agreement for $19$ of the $21$ latent positions.
Moreover, as the camera moves around the character, the dominant attention
response progressively shifts across anchors with different viewpoints,
showing that the memory branch retrieves character information according
to the evolving target camera geometry.

\paragraph{Effect of anchor geometry alignment on routing.}
We further compare the learning dynamics with and without Anchor Geometry
Alignment.
As shown in \redfigref{fig:anchor_routing_training}, geometry alignment
leads to faster improvement in routing accuracy during training and
achieves a higher final accuracy.
This suggests that explicitly relating anchor viewpoints to the target
camera geometry provides a stronger geometric prior for learning
view-aware memory retrieval.

\subsection{Additional Qualitative Results}

\begin{figure}[h]
	\centering
	\vspace{-5pt}
	\includegraphics[width=1.0\linewidth]{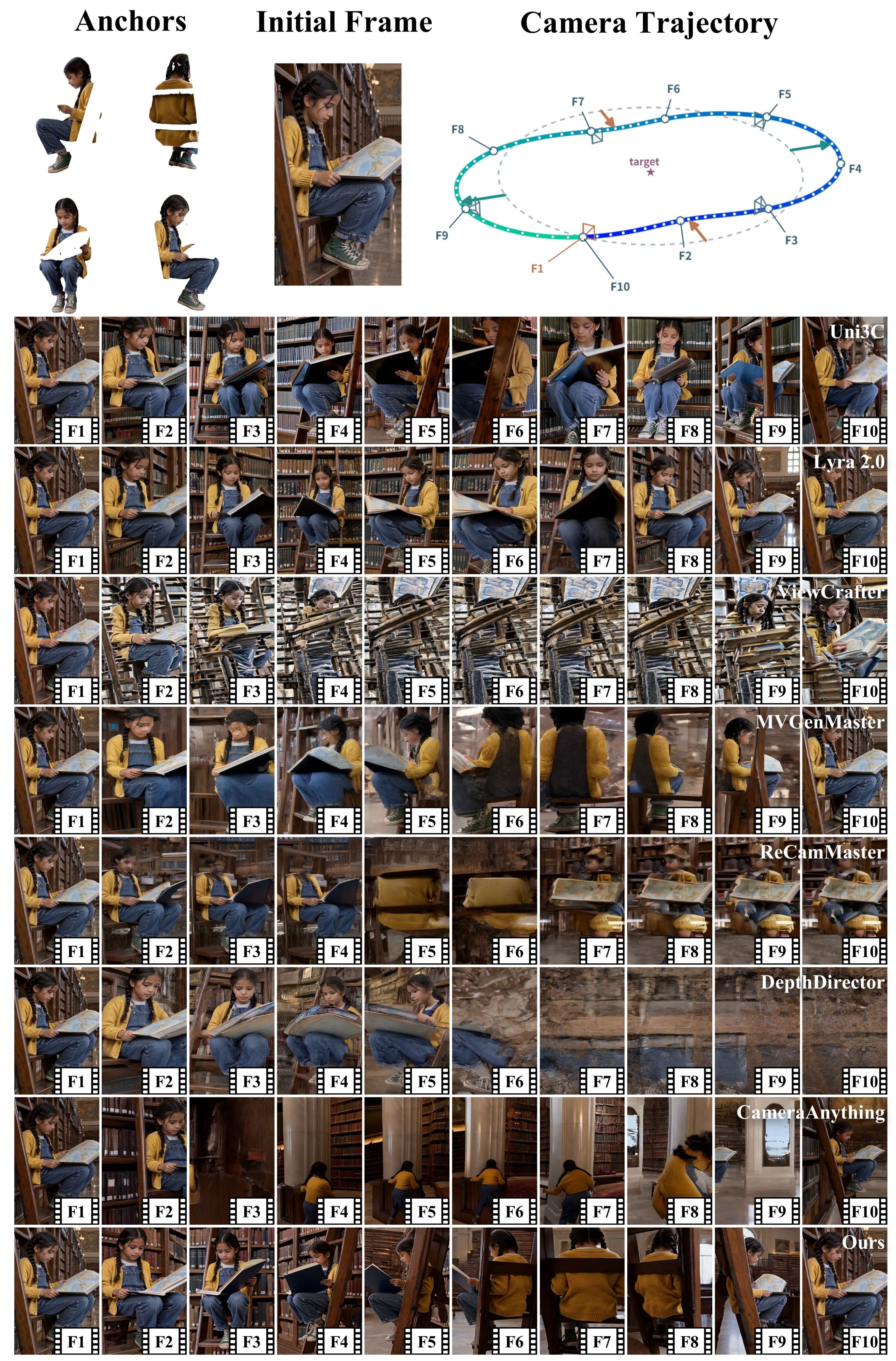}
	\vspace{-12pt}
	\caption{
		Additional qualitative comparison with state-of-the-art methods
		under diverse large-angle camera motions.
	}
	\label{fig:additional_qual_1}
	\vspace{-5pt}
\end{figure}

\begin{figure}[h]
	\centering
	\vspace{-5pt}
	\includegraphics[width=1.0\linewidth]{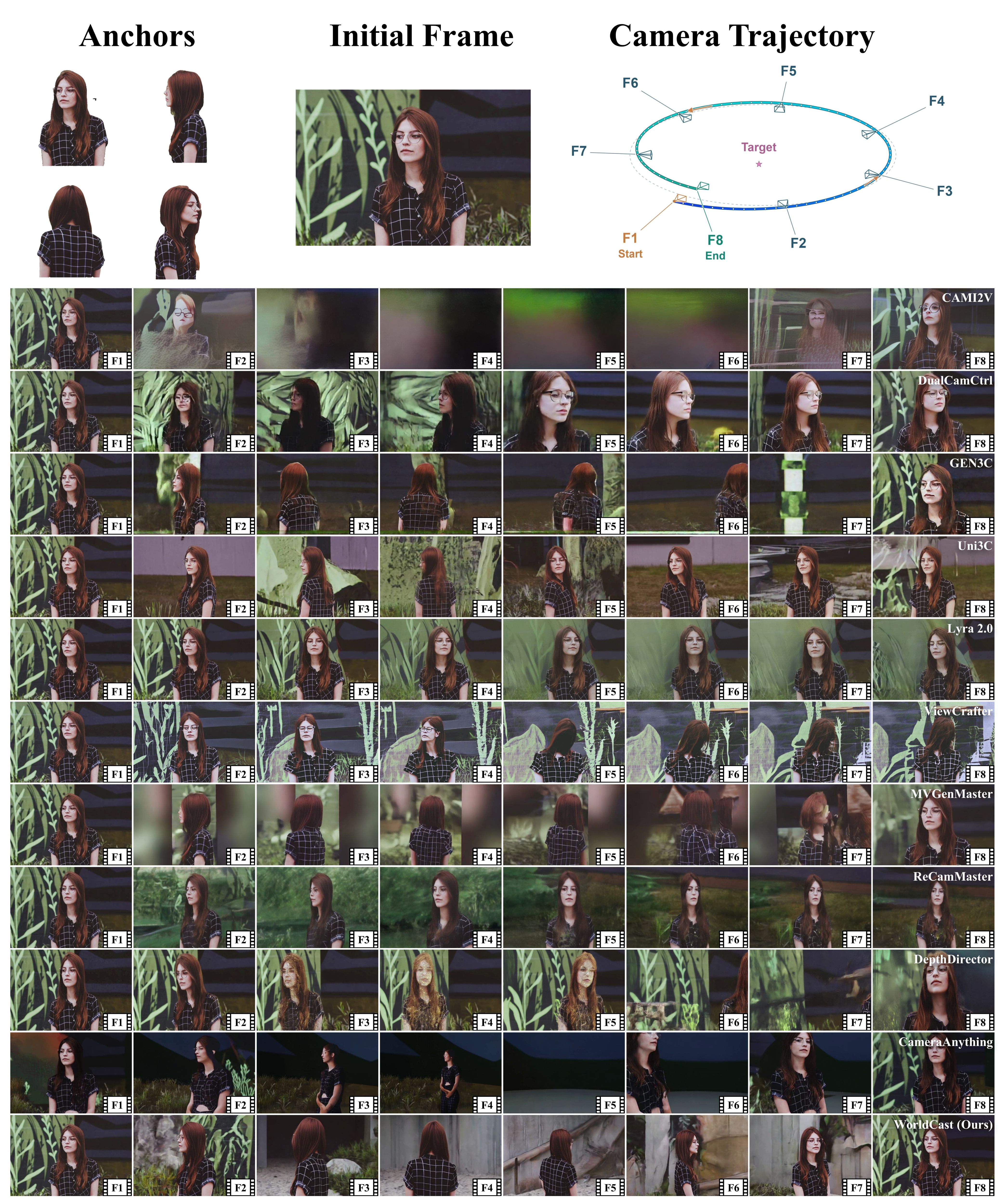}
	\vspace{-12pt}
	\caption{
		Additional qualitative comparison under large-angle camera motion
		with different initial viewpoints.
	}
	\label{fig:additional_qual_2}
	\vspace{-5pt}
\end{figure}

\begin{figure}[h]
	\centering
	\vspace{-5pt}
	\includegraphics[width=1.0\linewidth]{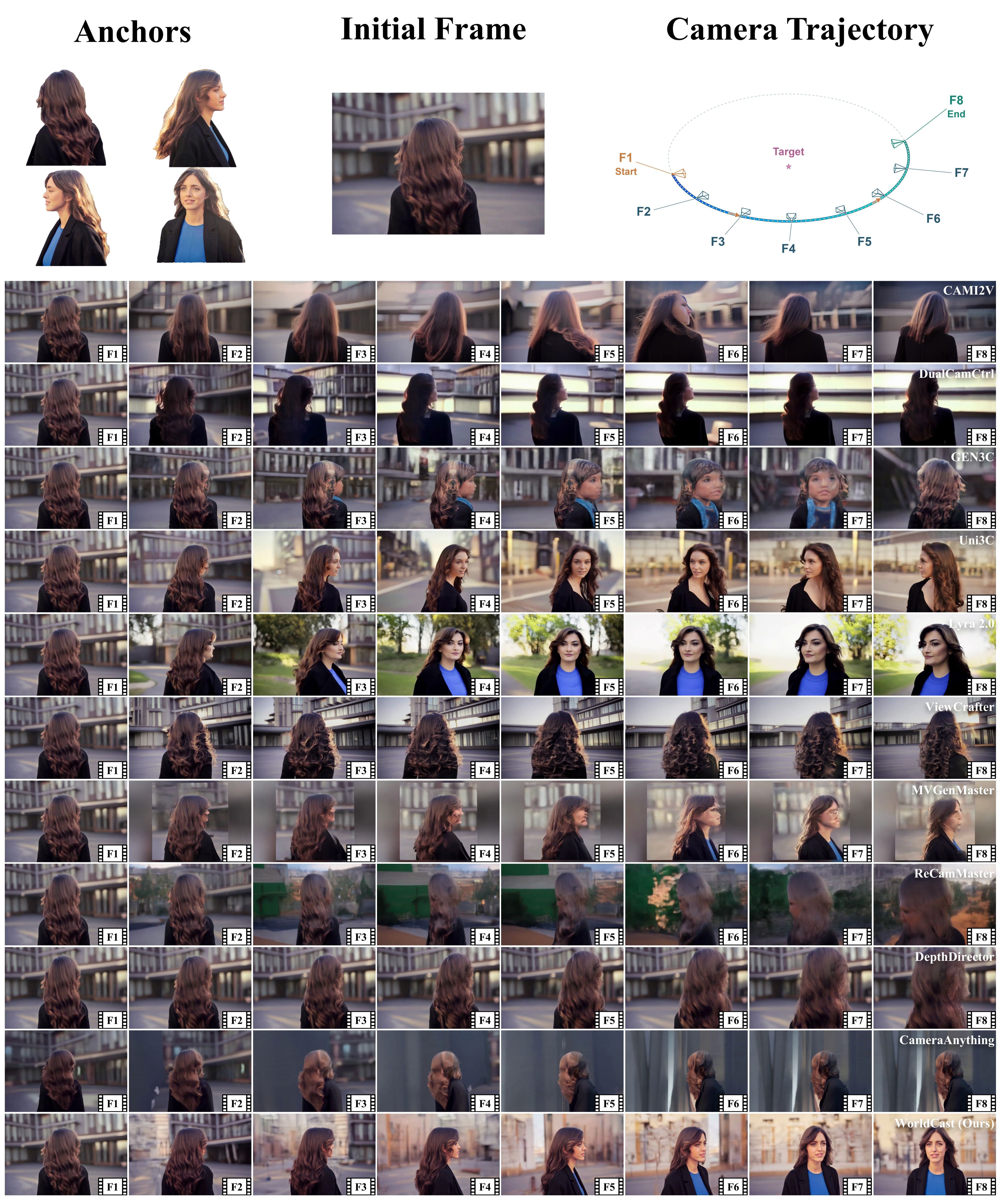}
	\vspace{-12pt}
	\caption{
		Additional qualitative comparison on combined rotation and
		translation trajectories.
	}
	\label{fig:additional_qual_3}
	\vspace{-5pt}
\end{figure}

\begin{figure}[h]
	\centering
	\vspace{-5pt}
	\includegraphics[width=0.90\linewidth]{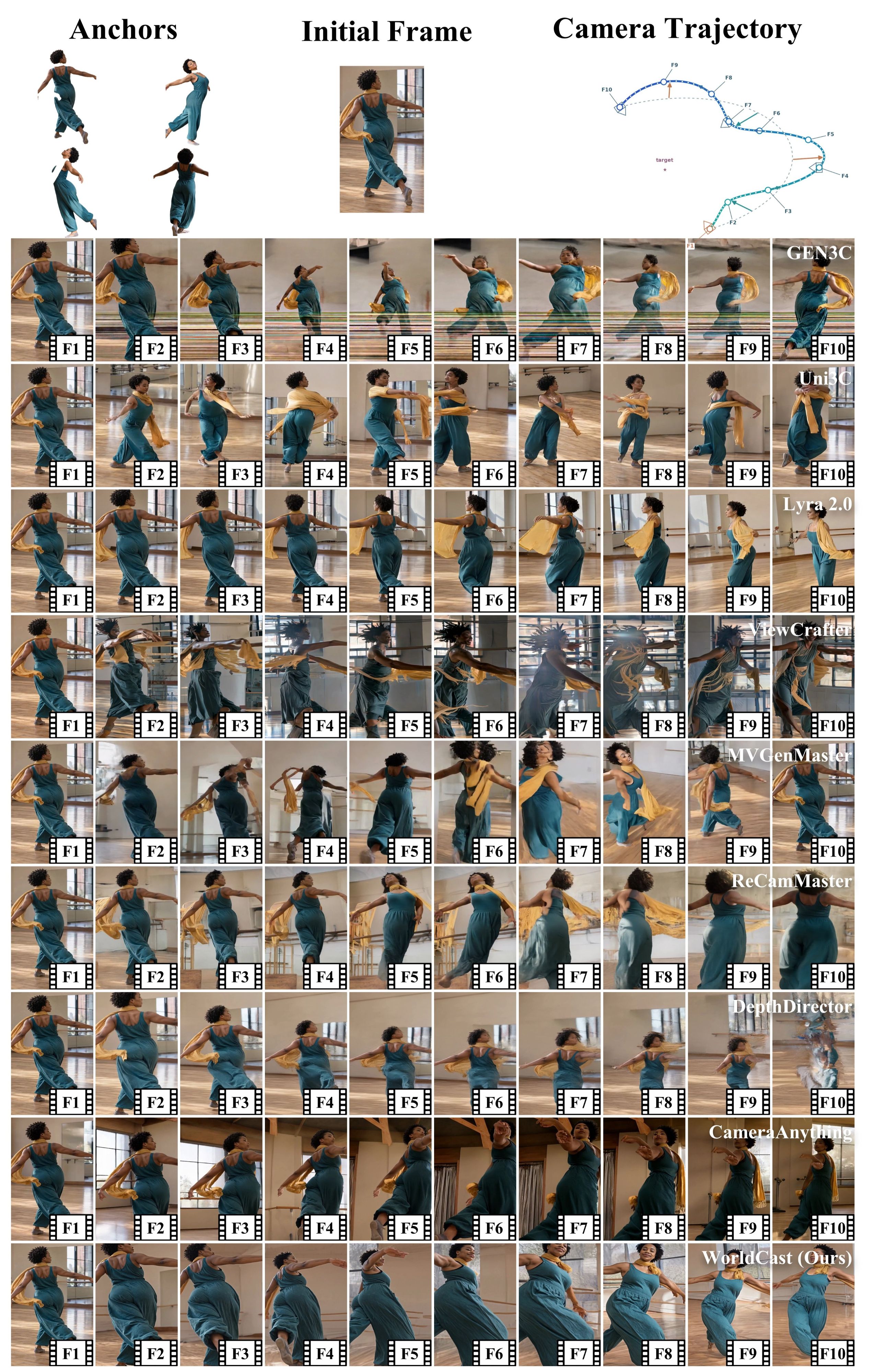}
	\vspace{-6pt}
	\caption{
		Additional qualitative comparison on complex large-angle
		camera trajectories.
	}
	\label{fig:additional_qual_4}
	\vspace{-5pt}
\end{figure}

\begin{figure}[h]
	\centering
	\vspace{-5pt}
	\includegraphics[width=0.93\linewidth]{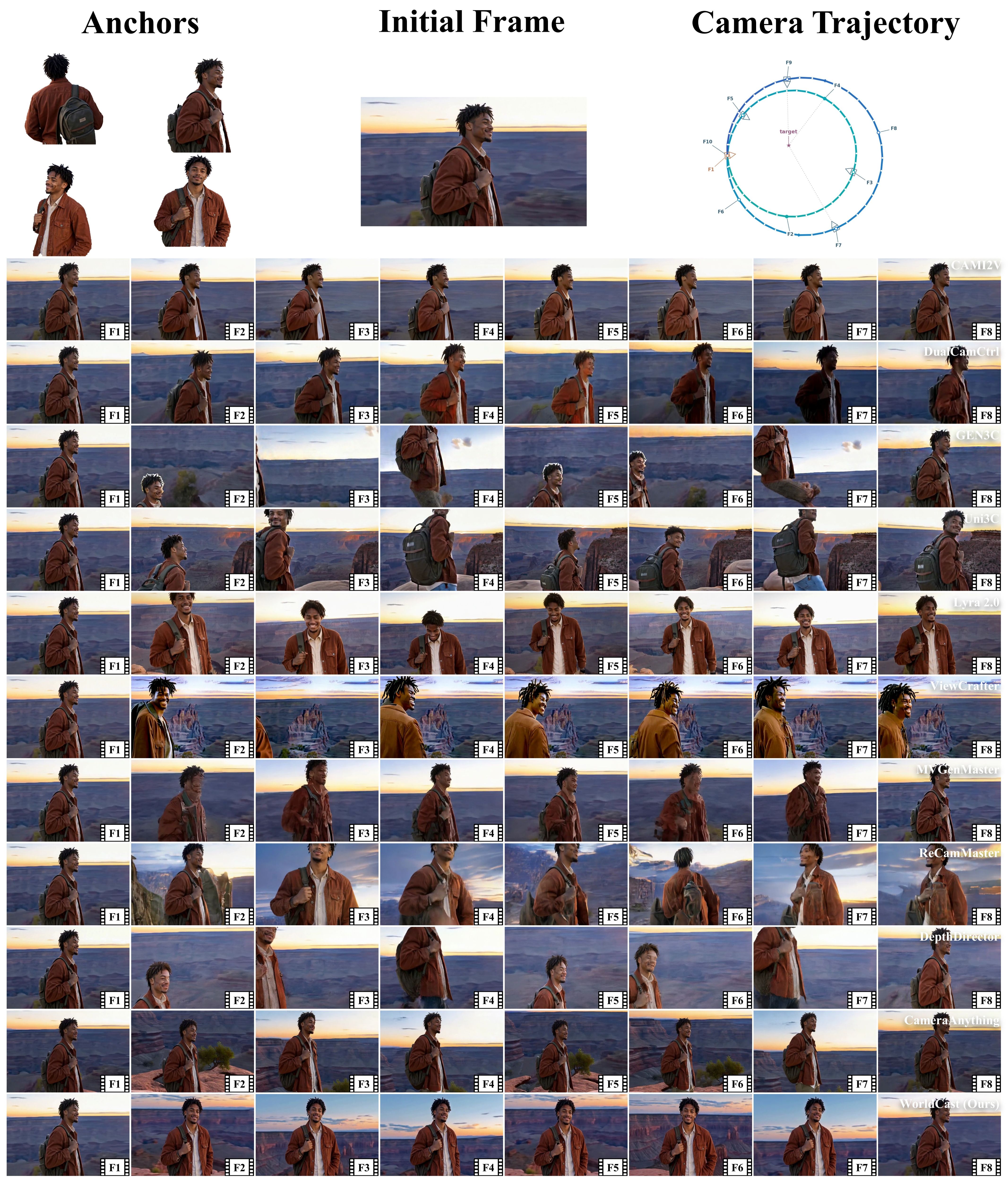}
	\vspace{-6pt}
	\caption{
		Additional qualitative comparison across diverse camera motions
		and initial azimuths.
	}
	\label{fig:additional_qual_5}
	\vspace{-5pt}
\end{figure}

\paragraph{Comparison across diverse trajectories.}
\redfigref{fig:additional_qual_1}--\redfigref{fig:additional_qual_5}
present additional comparisons across diverse camera trajectories,
including large-angle orbits, orbit-and-dolly motions, spiral and
hyperbolic paths, as well as conventional camera motions.
The examples also cover different initial azimuths, further evaluating
camera control and character preservation under substantial viewpoint
changes.

\begin{figure}[h]
	\centering
	\vspace{-5pt}
	\includegraphics[width=0.92\linewidth]{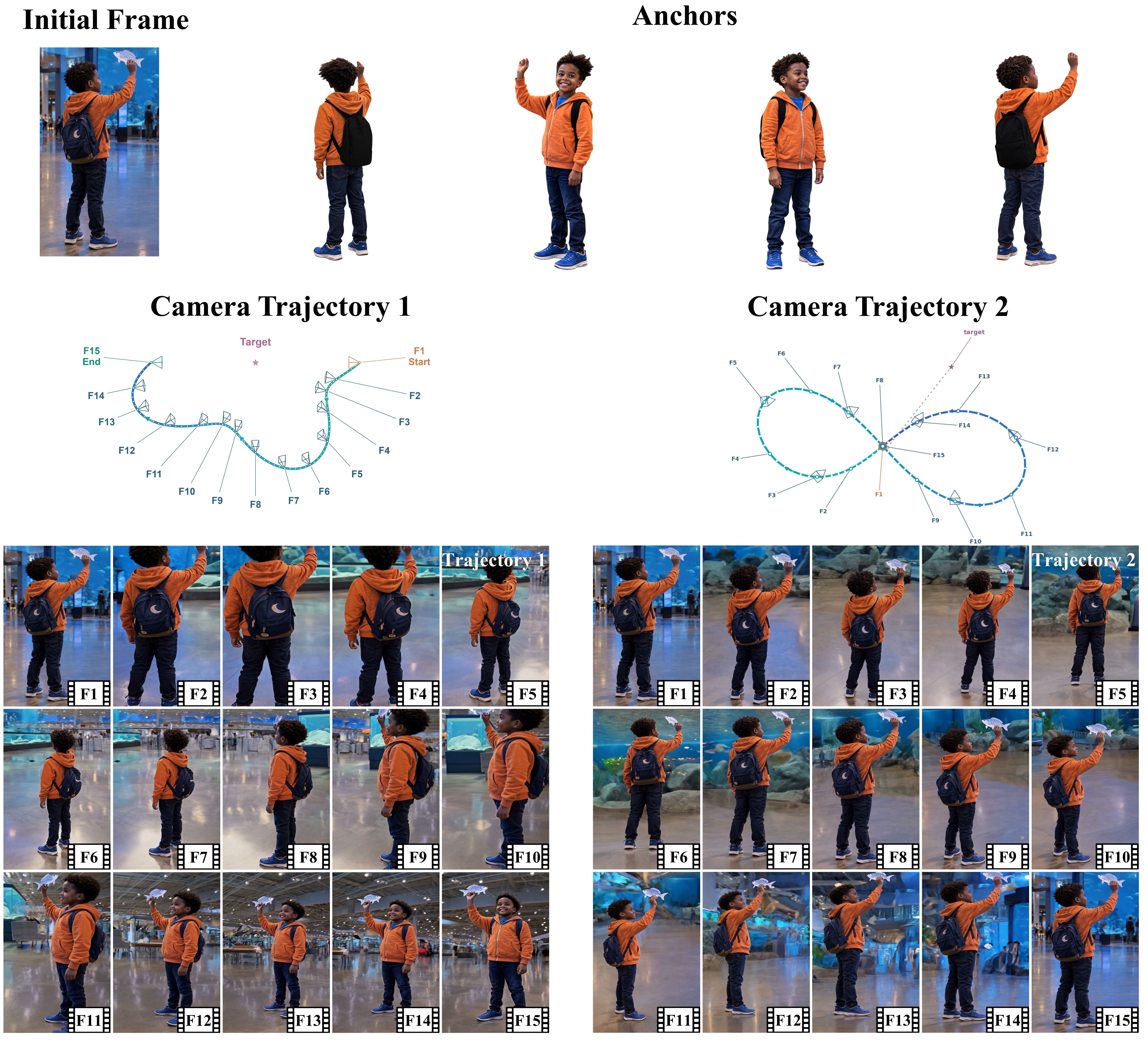}
	\vspace{-6pt}
	\caption{
		Trajectory generalization using the same set of character anchors
		across diverse target camera motions.
	}
	\label{fig:trajectory_generalization_1}
	\vspace{-5pt}
\end{figure}

\begin{figure}[h]
	\centering
	\vspace{-5pt}
	\includegraphics[width=0.92\linewidth]{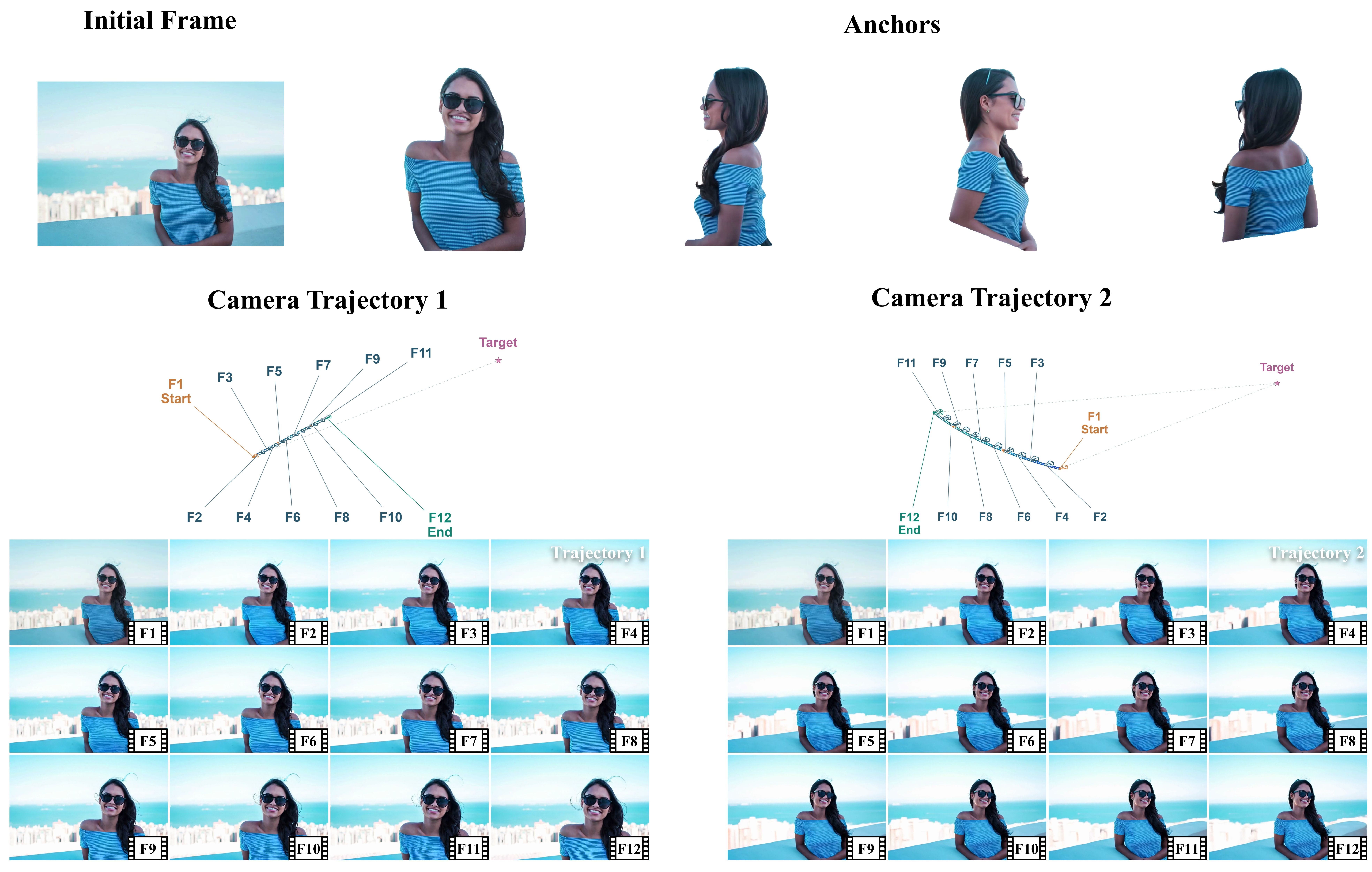}
	\vspace{-6pt}
	\caption{
		Trajectory generalization using the same set of character anchors
		across diverse target camera motions.
	}
	\label{fig:trajectory_generalization_2}
	\vspace{-5pt}
\end{figure}

\paragraph{Generalization across camera trajectories.}
We further evaluate trajectory generalization while keeping the same
set of character anchors fixed.
As shown in \redfigref{fig:trajectory_generalization_1} and
\redfigref{fig:trajectory_generalization_2}, the same multi-view
observations support diverse target camera motions, including orbits,
translations, and combined motions.
This shows that the character anchors are not tied to a particular
camera path and can be reused across different target trajectories.

\end{document}